\documentclass[sn-basic]{sn-jnl}

\usepackage{graphicx}%
\usepackage{multirow}%
\usepackage{amsmath,amssymb,amsfonts}%
\usepackage{amsthm}%
\usepackage{mathrsfs}%
\usepackage[title]{appendix}%
\usepackage{xcolor}%
\usepackage{textcomp}%
\usepackage{manyfoot}%
\usepackage{booktabs}%
\usepackage{algorithm}%
\usepackage{algorithmicx}%
\usepackage{algpseudocode}%
\usepackage{listings}%
\usepackage{caption}

\usepackage{tabularx,booktabs}
\usepackage{url}
\usepackage{fullpage}
\usepackage{hyperref}
\usepackage{amsmath}
\usepackage{amsfonts}
\usepackage{bbm}
\usepackage{color}
\usepackage{ulem}

\theoremstyle{thmstyleone}%
\theoremstyle{thmstyletwo}%

\theoremstyle{thmstylethree}%
\newtheorem{definition}{Definition}%

\newcommand{\ie}{\mbox{\it{i.e.,\ }}}
\newcommand{\eg}{\mbox{\it{e.g.,\ }}}
\newcommand{\quotes}[1]{``#1''}

\newcommand{\dperp}{\mathbin{\rotatebox[origin=c]{90}{$\models$}}}

\newcommand{\methodname}{Diamond}

\begin{document}

\title[Article Title]{Error-controlled non-additive interaction discovery in machine learning models}


\author[1]{\fnm{Winston} \sur{Chen}}\email{chenwt@umich.edu}
\equalcont{These authors contributed equally to this work.}

\author[2]{\fnm{Yifan} \sur{Jiang}}\email{yifan.jiang@uwaterloo.ca}
\equalcont{These authors contributed equally to this work.}

\author[3]{\fnm{William Stafford} \sur{Noble}}\email{william-noble@uw.edu}

\author*[2]{\fnm{Yang Young} \sur{Lu}}\email{yanglu@uwaterloo.ca}

\affil[1]{\orgdiv{Computer Science and Engineering Division}, \orgname{University of Michigan}, \city{Ann Arbor}, \state{Michigan}, \country{USA}}

\affil[2]{\orgdiv{Cheriton School of Computer Science}, \orgname{University of Waterloo}, \city{Waterloo}, \state{Ontario}, \country{Canada}}

\affil[3]{\orgdiv{Department of Genome Sciences and Paul G. Allen School of Computer Science and Engineering}, \orgname{University of Washington}, \city{Seattle}, \state{Washington}, \country{USA}}


\abstract{
Machine learning (ML) models are powerful tools for detecting complex patterns within data, yet their \quotes{black box} nature limits their interpretability, hindering their use in critical domains like healthcare and finance. To address this challenge, interpretable ML methods have been developed to explain how features influence model predictions. However, these methods often focus on univariate feature importance, overlooking the complex interactions between features that ML models are capable of capturing. Recognizing this limitation, recent efforts have aimed to extend these methods to discover feature interactions, but existing approaches struggle with robustness and error control, especially under data perturbations.

In this study, we introduce \methodname, a novel method for trustworthy feature interaction discovery. \methodname\ uniquely integrates the model-X knockoffs framework to control the false discovery rate (FDR), ensuring that the proportion of falsely discovered interactions remains low. 
A key innovation in \methodname\ is its non-additivity distillation procedure, which refines existing interaction importance measures to distill non-additive interaction effects, ensuring that FDR control is maintained. This approach addresses the limitations of off-the-shelf interaction measures, which, when used naively, can lead to inaccurate discoveries.
\methodname's applicability spans a wide range of ML models, including deep neural networks, transformer models, tree-based models, and factorization-based models. Our empirical evaluations on both simulated and real datasets across various biomedical studies demonstrate \methodname's utility in enabling more reliable data-driven scientific discoveries. This method represents a significant step forward in the deployment of ML models for scientific innovation and hypothesis generation.
}




\maketitle

\section{Introduction}
\label{sec:intro}

Machine learning (ML) has emerged as a critical tool in many application domains, largely due to its ability to detect subtle relationships and patterns within complex data \citep{obermeyer:predicting}.
While the complexity of ML models contributes to their power, it also makes them challenging to interpret, leaving users with few clues about the mechanisms underlying a given model's outputs.
Consequently, this \quotes{black box} nature of ML models, particularly deep neural networks, has hindered their applicability in error-intolerant domains like healthcare and finance. 
Stakeholders such as clinicians need to understand why and how the models make predictions before making important decisions, such as disease diagnosis \citep{lipton:mythos}.
Importantly, without providing insight into their internal mechanisms, ML models cannot be effectively used for making data-driven scientific discoveries, which are crucial for gaining human-understandable insights and driving successful innovation \citep{agrawal:artificial}. 

To enhance the interpretability of ML models for better data-driven scientific discoveries, interpretable ML methods have been developed to elucidate the internal mechanisms of these models \citep{samek:explaining}.
These methods help to elucidate how individual features influence prediction outcomes by assigning an importance score to each feature so that higher scores indicate greater relevance to the prediction \citep{simonyan:deep, shrikumar:learning, lundberg:unified, sundararajan:axiomatic, lu:dance}.
However, these univariate interpretations overlook a primary advantage of ML models: their ability to model complex interactions between features in a data-driven way.
In fact, input features usually do not work individually within an ML model but cooperate with other features to make inferences jointly \citep{tsang:detecting}.
For example, it is well established in biology that genes do not operate in isolation but work together in co-regulated pathways with additive, cooperative, or competitive interactions \citep{lu:wider}.
Additionally, gene-gene, gene-disease, gene-drug, and gene-environment interactions are critical in explaining genetic mechanisms, diseases, and drug effects \citep{watson:interpretable}.

Recognizing the limitations of univariate interpretations, efforts have been made to extend interpretable ML methods to discover interactions among features.
These methods attribute the prediction influence to feature pairs and then rank candidate feature pairs from a trained ML model, with highly ranked pairs indicating higher importance \citep{tsang:detecting, tsang:neural, cui:recovering, lundberg:local, janizek:explaining, sundararajan:shapley, chang:node-gam, lerman:explaining, zhang:interpreting}.
However, it is important to note that these approaches characterize feature pairs where both features are individually important for a model's prediction, rather than capturing the synergistic or interaction effects between the two features \citep{tsang:interpretable} (see the detailed explanation in Sec.\ref{sec:methods:distillation}).
Furthermore, the induced ranked list of feature pairs must be cut off at a certain confidence level for use in scientific discovery and hypothesis validation \citep{tsang:detecting}.
However, selecting this threshold is typically under user control and must be set arbitrarily.
Worse still, existing methods are sensitive to perturbations, in the sense that even imperceivable perturbations of the input data may lead to dramatic changes in the importance ranking \citep{ghorbani:interpretation, kindermans:unreliability, lu:ace}. 

From a practitioner's perspective, a given set of discovered feature interactions is scientifically valuable only if a systematic strategy exists to prioritize and select relevant interactions in a robust and error-controlled manner, even in the presence of noise.
Although many methods have been developed for feature interaction discovery, we are not aware of any previous attempts to conduct discovery while explicitly estimating and controlling the discovery error. Without this, accurate and reliable findings cannot be achieved.
In this study, we introduce an error-controlled interaction discovery method named \methodname\ (\underline{D}iscovering \underline{i}nter\underline{a}ctions in \underline{m}achine learning m\underline{o}dels with a co\underline{n}trolle\underline{d} error rate).
Here, the error is quantified by the false discovery rate (FDR) \citep{benjamini:controlling}, which informally represents the expected proportion of falsely discovered interactions among all discovered interactions. 
A false discovery is a feature interaction that is discovered but not truly relevant.

\begin{figure}[H]
\centering
\includegraphics[width=0.9\textwidth]{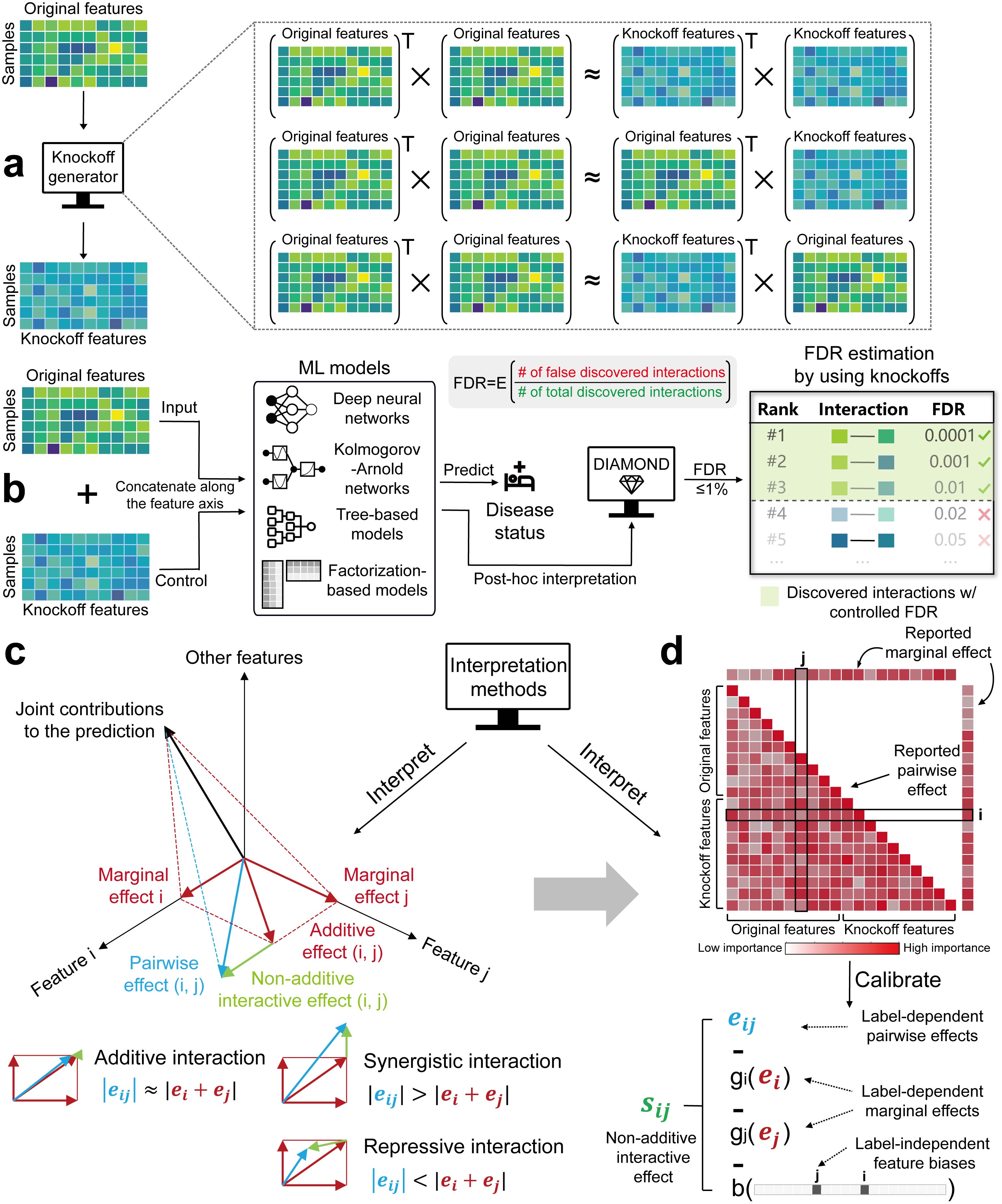}
\caption{
{\bf Overview of \methodname.}
(\textbf{a}) \methodname\ achieves false discovery rate (FDR) control by leveraging knockoffs -- dummy features that perfectly mimic the empirical dependence structure among the original features but are conditionally independent of the response given the original features.
(\textbf{b}) \methodname\ trains generic ML models using both the original features and their knockoff counterparts as inputs.
\methodname\ quantifies feature interactions from trained ML models and produces a ranked list of these interactions with estimated FDR, allowing users to confidently determine a cutoff threshold based on their desired confidence level.
(\textbf{c}) Existing feature importance measures are unable to directly capture non-additive interactions.
Geometrically, the marginal or interaction importance reported by current interpretation methods correspond to the projection of the total contribution to the prediction onto a one-dimensional feature axis or a two-dimensional feature-feature plane. 
Non-additive effects, however, represent the difference between the interaction effects and the marginal effects.
These non-additive effects can manifest as either synergistic or repressive interactions.
(\textbf{d}) \methodname\ distills non-additive effects from the interaction importance measures reported by existing methods, thereby maintaining FDR control at the target level.
The distillation procedure is designed to remove both label-dependent marginal effects and label-independent feature biases from the reported feature interactions, leaving only the label-dependent non-additive interaction effects.
\label{fig:overview}}
\end{figure}

Three components of \methodname\ are novel (Fig. \ref{fig:overview}).
First, \methodname\ achieves FDR control by leveraging the model-X knockoffs framework \citep{barber:controlling, candes:panning}. 
The core idea of this framework is to generate dummy features that perfectly mimic the empirical dependence structure among the original features while being conditionally independent of the response given the original features.
Second, we discover that naively using off-the-shelf feature interaction importance measures cannot correctly control the FDR. 

To address this issue, we distill non-additive interaction effects from the interaction importance measures reported by existing methods, ensuring FDR control at the target level.
Third, \methodname\ is applicable to a wide range of ML models, including deep neural networks (DNNs), Kolmogorov-Arnold networks (KANs) \citep{liu:kan}, transformer models \citep{gorishniy:revisiting}, tree-based models, and factorization-based models.

We have applied \methodname\ to various simulated and real datasets to demonstrate its empirical utility.
Practically speaking, \methodname\ paves the way for the wider deployment of machine learning models in scientific discovery and hypothesis generation.

\section{Results}
\label{sec:results}

\subsection{\methodname\ overview}
\label{sec:results:overview}

Consider a prediction task where we have $n$ independent and identically distributed (i.i.d.) samples $\mathbf{X}=\left \{ x_i \right \}_{i=1}^{n}  \in \mathbb{R}^{n \times p}$ and $\mathbf{Y}=\left \{ y_i \right \}_{i=1}^{n} \in \mathbb{R}^{n \times 1}$, denoting the data matrix with $p$-dimensional features and the corresponding response, respectively.
The response of the prediction task can be categorical labels such as disease status or numerical measurements such as body mass index, and the features could include gene expression data, microbial taxa abundance profiles, and more.
The prediction task can be described using an ML model $f: \mathbb{R}^{p} \mapsto \mathbb{R}$ that maps from the input $x \in \mathbb{R}^{p}$ to the response $y \in \mathbb{R}$.
When modeling the task, the function $f$ learns the dependence structure of $\mathbf{Y}$ on $\mathbf{X}$, enabling effective prediction with the fitted ML model.

In this study, we focus on interpreting the fitted ML model for data-driven scientific discovery by identifying the non-additive feature interactions that contribute most to the model's predictions.
We say that $\mathcal{I} \subset \{1,\cdots, p\}$ is a non-additive interaction of function $f$ if and only if $f$ cannot be decomposed into an addition of $\left | \mathcal{I} \right |$ subfunctions $f_i$, each of which excludes a corresponding interaction feature \citep{sorokina:detecting,tsang:detecting}, \ie $f(x)\neq \sum_{i \in \mathcal{I}}f_i\left ( x_{\{1,\cdots, p\}\backslash i} \right )$.
For example, the multiplication between two features $x_i$ and $x_j$ is a non-additive interaction because it cannot be decomposed into a sum of univariate functions, \ie $f(x_i,x_j)=x_ix_j\neq f_i(x_j)+f_j(x_i)$.
On the other hand, $\log (x_i x_j)$ is considered to be an additive interaction because it can be decomposed in the logarithmic space, \ie $f(x_i,x_j)=\log (x_i x_j)= \log (x_j)+\log(x_i)$. 
Assume that there exists a group of interactions $\mathcal{S}=\left \{ \mathcal{I}_1,\mathcal{I}_2,\cdots \right \}$ such that, 
conditional on interactions $\mathcal{S}$, the response $\mathbf{Y}$ is independent of interactions in the complement $\mathcal{S}^c=\{1,\cdots, p\}\times \{1,\cdots, p\}\backslash \mathcal{S}$.
In this setting, our goal is to accurately discover feature interactions in $\mathcal{S}$ without erroneously reporting too many incorrect interactions in $\mathcal{S}^c$. 

\methodname\ is designed to discover feature interactions from fitted ML models while maintaining a controlled FDR \citep{benjamini:controlling}.
For a set of feature interactions $\widehat{S} \subset \{1,\cdots, p\}\times \{1,\cdots, p\}$ discovered by some interaction detection method, the FDR is defined as:
\begin{align*}
\text{FDR} = \mathbb{E}[\text{FDP}] \text{ with } \text{FDP} = \frac{|\widehat{S}\cap \mathcal{S}^c|}{|\widehat{S}|}.
\end{align*}
Commonly used procedures, such as the Benjamini--Hochberg procedure \citep{benjamini:controlling}, achieve FDR control by working with p-values computed against some null hypothesis.
In the interaction discovery setting, for each feature interaction, one tests the significance of the statistical association between the specific interaction and the response, either jointly or marginally, and obtains a p-value under the null hypothesis that the interaction is irrelevant.
These p-values are then used to rank the features for FDR control.
However, controlling FDR in generic machine learning models, especially deep learning models, is challenging because, to our knowledge, the field lacks a method for producing meaningful p-values that reflect interaction importance.
To bypass the use of p-values but still achieve FDR control, we draw inspiration from the model-X knockoffs framework \citep{barber:controlling, candes:panning}.
The core idea of knockoffs is to generate dummy features that perfectly mimic the empirical dependence structure among the original features but are conditionally independent of the response given the original features.
These knockoff features can then be used as a control by comparing the feature importance between the original features and their knockoff counterparts to estimate FDR (Fig. \ref{fig:overview}\textbf{a}). 

\methodname\ trains a generic ML model that takes as input an augmented data matrix $(\mathbf{X}, \tilde{\mathbf{X}}) \in \mathbb{R}^{n \times 2p}$, created by concatenating the data matrix $\mathbf{X} \in \mathbb{R}^{n \times p}$ with its knockoffs $\tilde{\mathbf{X}} \in \mathbb{R}^{n \times p}$ along the feature axis.
After training, \methodname\ quantifies feature interactions by interpreting the trained model and produces a ranked list of these interactions (Fig. \ref{fig:overview}\textbf{b}).
However, existing feature importance measures cannot directly capture non-additive interactions.
Geometrically, the marginal or interaction importance reported by current interpretation methods correspond to the projection of the total contribution to the prediction onto a one-dimensional feature axis or a two-dimensional feature-feature plane. 
Non-additive effects, however, represent the difference between the interaction effects and the marginal effects (Fig. \ref{fig:overview}\textbf{c}).
Hence, \methodname\ distills non-additive interaction effects from the reported interaction importance measures from existing methods, thereby maintaining FDR control at the target level.
The distillation process is designed to remove both label-dependent marginal effects and label-independent feature biases from the reported feature interactions, leaving only the label-dependent non-additive interaction effects (Fig. \ref{fig:overview}\textbf{d}).

\subsection{\methodname\ discovers FDR-controlled interactions on simulated datasets}
\label{sec:result:simulation}
\begin{figure}[H]
\centering
\includegraphics[width=1.0\textwidth]{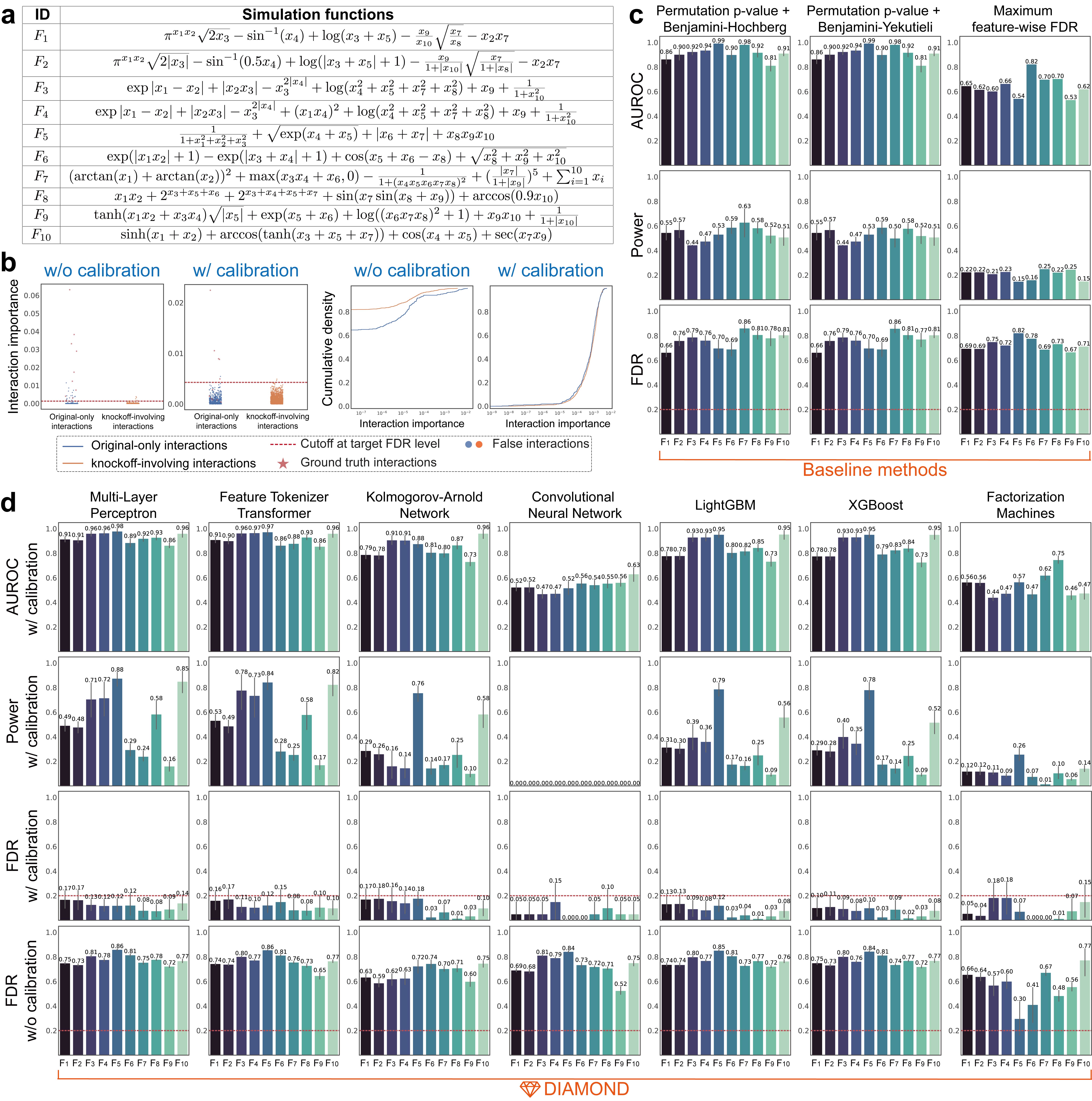}
\caption{
{\bf Evaluating \methodname\ for FDR control on simulated datasets.}
(\textbf{a}) The evaluation is based on a test suite of 10 data-generating simulation functions proposed by \cite{tsang:detecting}.
(\textbf{b}) The reported interaction importance from existing methods in simulation function $F_1$ reveals a clear distribution disparity between original-only interactions and those involving knockoffs.
The distilled non-additive interactions help mitigate distributional disparities.
(\textbf{c}) Baseline methods fail to correctly control the FDR, thereby rendering the reported high power and AUROC invalid.
(\textbf{d}) \methodname\ identifies important non-additive interactions with controlled FDR, compatible with various ML models.
The non-additivity distillation procedure is critical; without it, the FDR cannot be controlled.
Even in scenarios of model misspecification, such as using convolutional neural networks for tabular data, \methodname\ maintains FDR control, albeit with a loss of power.
\label{fig:simulation}}
\end{figure}

We started by evaluating the performance of \methodname\ on simulated datasets, assessing its ability to identify important non-additive interactions while controlling the FDR.
We benchmarked \methodname\ on a test suite of 10 simulated datasets generated by different simulation functions proposed by \cite{tsang:detecting}. 
These datasets contain a mixture of univariate functions and multivariate interactions, exhibiting varied order, strength, and nonlinearity (Fig. \ref{fig:simulation}\textbf{a}).
Since our goal is to detect pairwise interactions, high-order interaction functions (\eg $F(x_1,x_2,x_3)=x_1 x_2 x_3$) are decomposed into pairwise interactions (\eg $(x_1,x_2)$, $(x_1,x_3)$, and  $(x_2,x_3)$) to serve as the ground truth.

Following the settings used in \cite{tsang:detecting}, we employed a sample size of $n=20,000$, equally divided into training and test sets. 
In addition, the number of features is set at $p=30$, and all features are sampled randomly from a continuous uniform distribution, $U(0,1)$.
Only the first $10$ out of $30$ features contribute to the corresponding response, while the remaining features serve as noise to increase the task's complexity.
For robustness, we repeated the experiment 20 times for each simulated dataset using different random seeds. 
Each repetition involved data generation, knockoff generation using KnockoffsDiagnostics \citep{blain:when}, ML model training, and interaction-wise FDR estimation.
For all simulation settings, we reported the mean performance with 95\% confidence intervals, fixing the target FDR level at $q=0.2$.

Our analysis shows that \methodname\ consistently identifies important non-additive interactions with controlled FDR across all ML models (Fig. \ref{fig:simulation}\textbf{d}).
Under controlled FDR, the multi-layer perceptron (MLP) and Feature Tokenizer Transformer (FT-Transformer) models \citep{gorishniy:revisiting} demonstrate better performance in identifying important interactions than other ML models, as measured by the statistical power and the area under the receiver operating characteristic curve (AUROC).
This superior performance is attributed to the inclusion of a knockoff-tailored plugin pairwise-coupling layer design \citep{lu:deeppink}, which has been shown to maximize statistical power (See details in Sec. \ref{sec:supp:mlp}).
It is important to note that certain ML models, such as convolutional neural networks, are not well-suited for modeling tabular data due to their design for capturing local patterns in images or text.
Even in the presence of model misspecification, \methodname\ maintains FDR control, albeit with a loss of power.
This highlights \methodname's robustness and its broad applicability across various ML models.

We discovered that the proposed non-additivity distillation (Sec \ref{sec:methods:distillation}) is critical; without it, the FDR cannot be controlled by naively using reported interaction importance values from existing methods (Fig. \ref{fig:simulation}\textbf{d} and Fig. \ref{fig:supp:simulation_calib}).
To gain insight into the FDR control failure in the absence of non-additivity distillation, we conducted a qualitative comparison assessing interaction importance before and after distillation using the simulation function $F_1$ (Fig. \ref{fig:simulation}\textbf{b} and Fig. \ref{fig:supp:simulation_distillation}). 
The primary cause of the FDR control failure appears to lie in the distributional disparity between original-only interactions (\ie interactions involving two features from the original feature set rather than  knockoffs) and interactions involving knockoffs.
This observation suggests a violation of the knockoff filter's assumption in controlling the FDR (See discussion in Sec \ref{sec:methods:distillation}).
The proposed non-additivity distillation procedure mitigates the disparity by extracting non-additive interaction effects from the reported interaction importance measures, thereby enhancing the utility of knockoff-involving interactions as a negative control for FDR estimation.

Finally, we verified whether alternative baseline methods can accurately identify important non-additive interactions with controlled FDR (Fig. \ref{fig:simulation}\textbf{c}).
We compared three baseline methods for FDR estimation: two relying on permutation-based interaction-wise p-values, combined with the Benjamini-Hochberg and the Benjamini-Yekutieli procedures respectively, and one that represents interaction-wise FDR as an aggregation of feature-wise FDR (See details in Sec. \ref{sec:methods:baselines}).
Our analysis reveals that none of these baseline methods effectively control the FDR.
This greatly reduces the utility of these methods, despite their reported high power and AUROC.

\subsection{\methodname\ is robust to various knockoffs and importance measures}
\label{sec:result:robust}
\begin{figure}[H]
\centering
\includegraphics[width=1.0\textwidth]{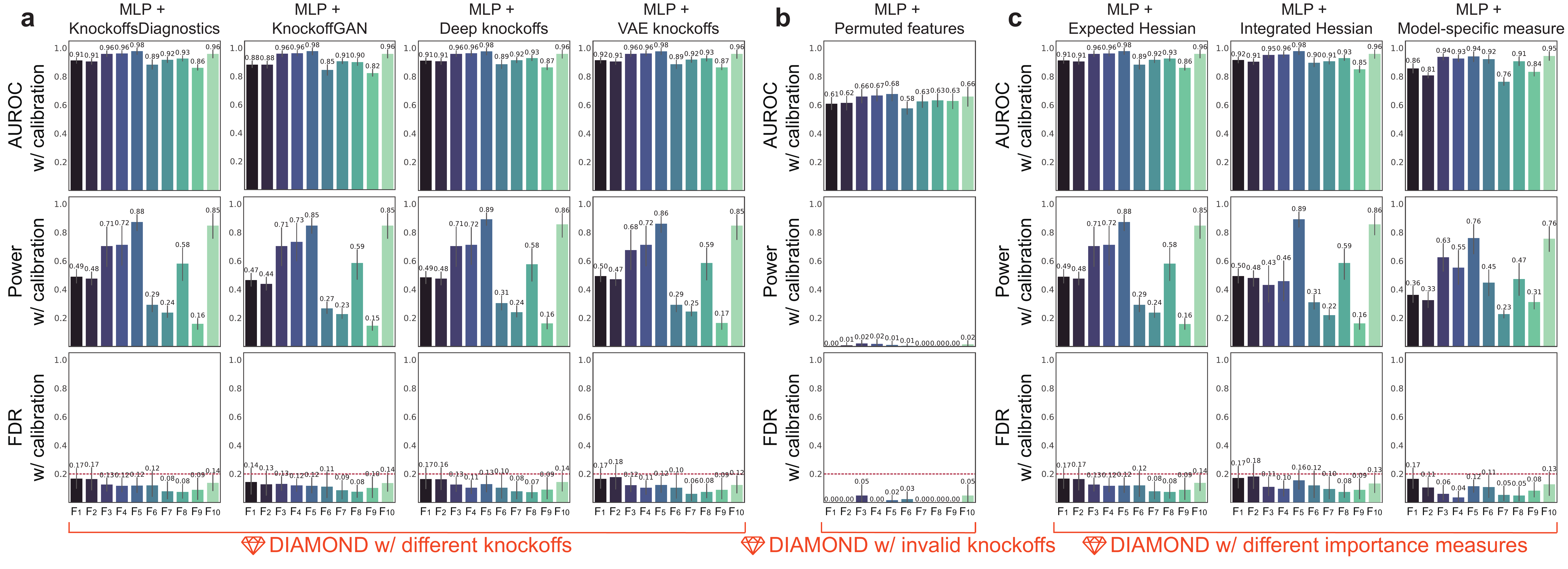}
\caption{
{\bf Evaluating \methodname\ for robustness on simulated datasets.}
(\textbf{a}) \methodname\ demonstrates robustness across knockoffs generated by various methods, including KnockoffsDiagnostics, KnockoffGAN, Deep knockoffs, and VAE knockoffs.
(\textbf{b}) \methodname\ maintains FDR control when paired with invalid knockoffs, generated by independently permuting each feature across samples, albeit at the cost of reduced power.
(\textbf{c}) \methodname\ demonstrates robustness across methodologically different interaction importance measures: Expected Hessian, Integrated Hessian, and model-specific measures.
\label{fig:robust}}
\end{figure}

The design of \methodname\ incorporates several key components, including knockoff generation and interaction importance measures.
To demonstrate the robustness of \methodname\ in FDR control, we conducted a control study where we modified \methodname\ by replacing these two components with alternative solutions.
Specifically, we considered two variants of \methodname\ using the MLP model.
In the first setting, we replaced KnockoffsDiagnostics with three alternatives: KnockoffGAN \citep{jordon:knockoffgan}, Deep knockoffs \citep{romano:deep} or VAE knockoffs \citep{liu:auto-encoding}, while keeping the MLP model and the interaction importance measure (\ie Expected Hessian \citep{erion:improving}) unchanged.
In the second setting, we replaced the interaction importance measure, Expected Hessian, with two alternatives: Integrated Hessian \citep{janizek:explaining} and a model-specific interaction importance measure derived from the MLP weights (See details in Sec. \ref{sec:supp:model-specific}), while keeping the MLP model and the knockoff generation unchanged.
For each variant, we applied \methodname\ to the test suite of 10 simulated datasets using the same settings. 

The results indicate that \methodname\ is robust to various knockoff designs and importance measures.
In the first study, our analysis shows that \methodname\ consistently discovers important non-additive interactions with controlled FDR across various knockoff designs (Fig. \ref{fig:robust}\textbf{a}).
Despite the methodological differences in knockoff generation between KnockoffsDiagnostics, KnockoffGAN, Deep knockoffs, and VAE knockoffs, \methodname\ achieves comparable statistical power and AUROC in each case. 
This result demonstrates the \methodname's stability in identifying important interactions in knockoff-based FDR estimation.
Furthermore, it is worth noteworthy that \methodname\ is robust even when the knockoffs are poorly generated.
Specifically, we generated invalid knockoffs by independently permuting each feature across samples, thereby violating the definition of knockoffs outlined in Sec. \ref{sec:methods:knockoffs}.
Our analysis demonstrates that even in this extreme scenario, \methodname\ maintains FDR control, albeit with reduced power (Fig. \ref{fig:robust}\textbf{b}).

In the second study, our analysis shows that \methodname\ robustly identifies FDR-controlled non-additive interactions across a range of interaction importance measures (Fig. \ref{fig:robust}\textbf{c}).
Despite the methodological differences between Expected Hessian, Integrated Hessian, and model-specific measures, which involve mechanistic disparities in elucidating the relationships between features and responses, \methodname\ achieves valid FDR control while maintaining reasonably good statistical power and AUROC.
This result demonstrates \methodname's flexibility in identifying interactions based on various importance measures.

Finally, it is worth mentioning that regardless of the knockoff generation and interaction importance measures we used, there is room for improvement in achieving better statistical power while maintaining controlled FDR.
\methodname\ tends to be conservative by overestimating the FDR, suggesting that an improved FDR control procedure could potentially boost statistical power.

\subsection{\methodname\ identifies interactions to explain disease progression}
\label{sec:result:diabetes}
\begin{figure}[H]
\centering
\includegraphics[width=1.0\textwidth]{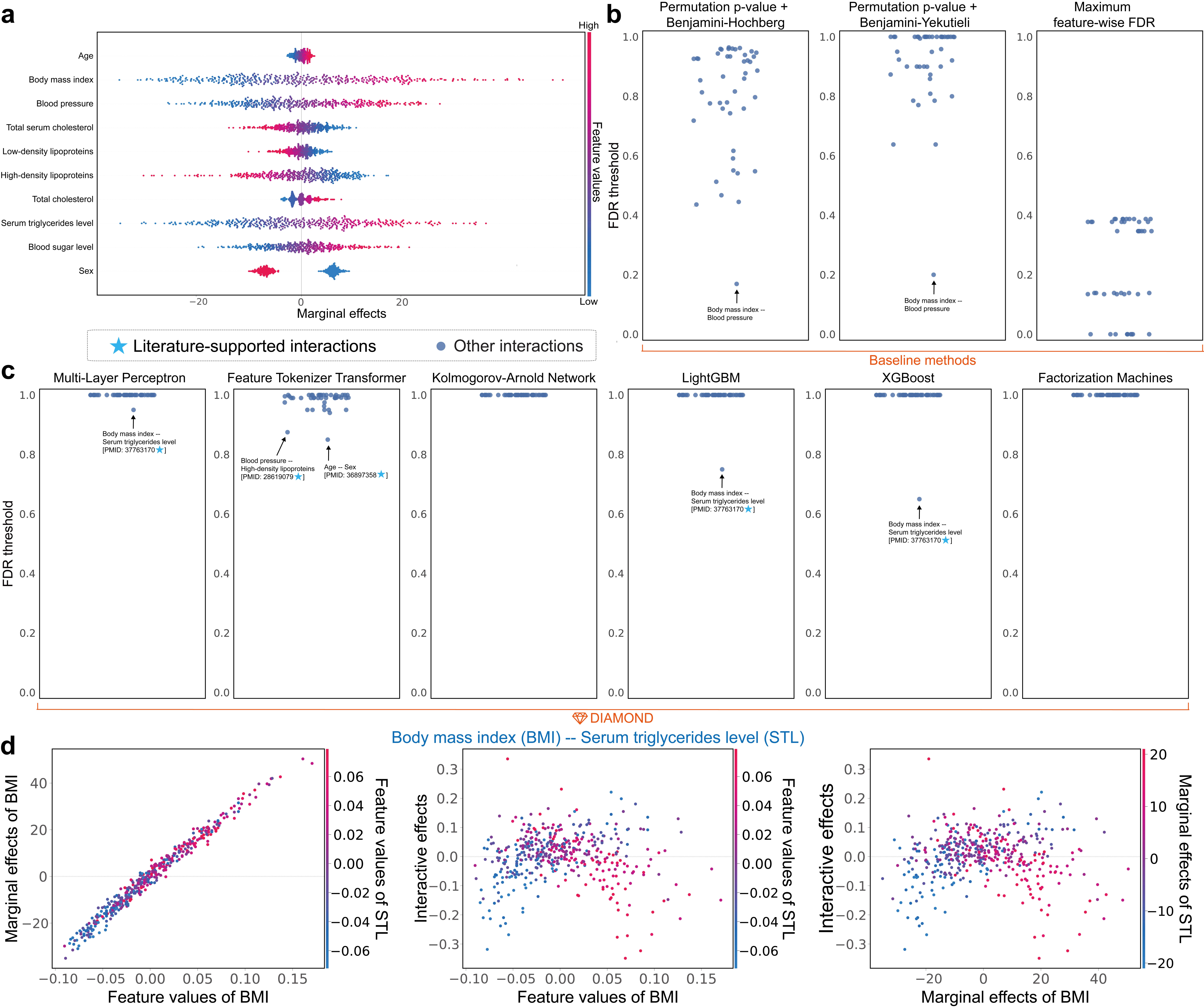}
\caption{
{\bf Evaluating \methodname\ on a real diabetes dataset.}
(\textbf{a}) Each feature contributes differently to predicting disease progression, as measured by the Expected Gradient scores in the MLP model.
(\textbf{b}) \methodname\ is compared against three baseline methods.
The blue stars indicate interactions supported by literature evidence, referenced by the accompanying PubMed identifiers.
(\textbf{c}) \methodname\ is used with various ML models to identify important non-additive interactions.
Each possible interaction is measured by the minimum FDR threshold cutoff at which it is selected, with the top interaction annotated.
It is worth mentioning that two marginally important features do not necessarily result in important interactions, as anticipated in \methodname's design.
(\textbf{d}) The top interaction, between body mass index and serum triglycerides level, is qualitatively evaluated from three aspects: the marginal importance measure, the interaction importance measure, and the contribution of the marginal importance measures to the interaction importance measure.
\label{fig:diabetes}}
\end{figure}
We then evaluated the performance of \methodname\ on a real dataset \citep{efron:least}, assessing its ability to identify important interactions in the progression of disease. 
We used a quantitative study with $n=442$ diabetes patients, each characterized by $10$ standardized baseline features, including age, sex, body mass index, average blood pressure, and six blood serum measurements (Fig. \ref{fig:diabetes}\textbf{a}).
The task is to construct an ML model to predict the response of interest, which is a quantitative measure of disease progression one year after the baseline.
We considered different types of ML models: MLP, FT-Transformer, KAN, LightGBM, XGBoost, and factorization machines (FM).
We assessed the trained ML model using 5-fold cross-validation and selected the one with the best performance for model interpretation and interaction discovery.
For robustness, we repeated each experiment 20 times using different random seeds. 
Each repetition involved knockoff generation, ML model training, and interaction-wise FDR estimation.

Our analysis shows that \methodname\ identifies the important interaction between body mass index (BMI) and serum triglycerides level (STL) across three different ML models (Fig. \ref{fig:diabetes}\textbf{c}). 
We investigated the identified interaction via literature search and found that high BMI and high STL are associated with an increased risk of diabetes, as reported by multiple studies \citep{garbuzova:triglycerides, zhao:high, huang:lower}.
The literature evidence is further supported by the qualitative evaluation (Fig. \ref{fig:diabetes}\textbf{d}).
Specifically, we examined the identified interactions from three perspectives: the marginal importance measure, the interaction importance measure, and the contribution of the marginal importance measures to the interaction importance measure.
This analysis showed that high BMI and high STL contribute to more severe progression of diabetes.
Meanwhile, higher STL further reinforces the contribution of BMI to the progression of diabetes synergistically.

Unlike other ML models, the FT-Transformer identifies two distinct interactions: the interaction between blood pressure and high-density lipoproteins, and the interaction between age and sex.
We investigated the identified interactions through a literature search and found that high-density lipoprotein levels are positively associated with hypertension in individuals with elevated levels of circulating CD34-positive cells \citep{shimizu:association}.
We also found that the prevalence of diabetes varies by sex, with men typically being diagnosed at a younger age and lower BMI, while young women with diabetes are currently less likely than men to receive treatment \citep{kautzky:sex}.

Finally, we find that \methodname\ does not necessarily report a lower FDR for interactions composed of two marginally important features.
All $10$ baseline features from diabetes patients are effective in predicting disease progression, as measured by the Expected Gradient scores in the MLP model (Fig. \ref{fig:diabetes}\textbf{a}).
Among these features, in addition to BMI and STL, blood pressure, blood sugar level, and high-density lipoproteins are more predictive than others.
It is important to note that \methodname\ still associates most interactions composed of these highly predictive features with a high FDR threshold.

\subsection{\methodname\ investigates enhancer activity in \textit{Drosophila} embryos}
\label{sec:result:enhancer}
\begin{figure}[H]
\centering
\includegraphics[width=1.0\textwidth]{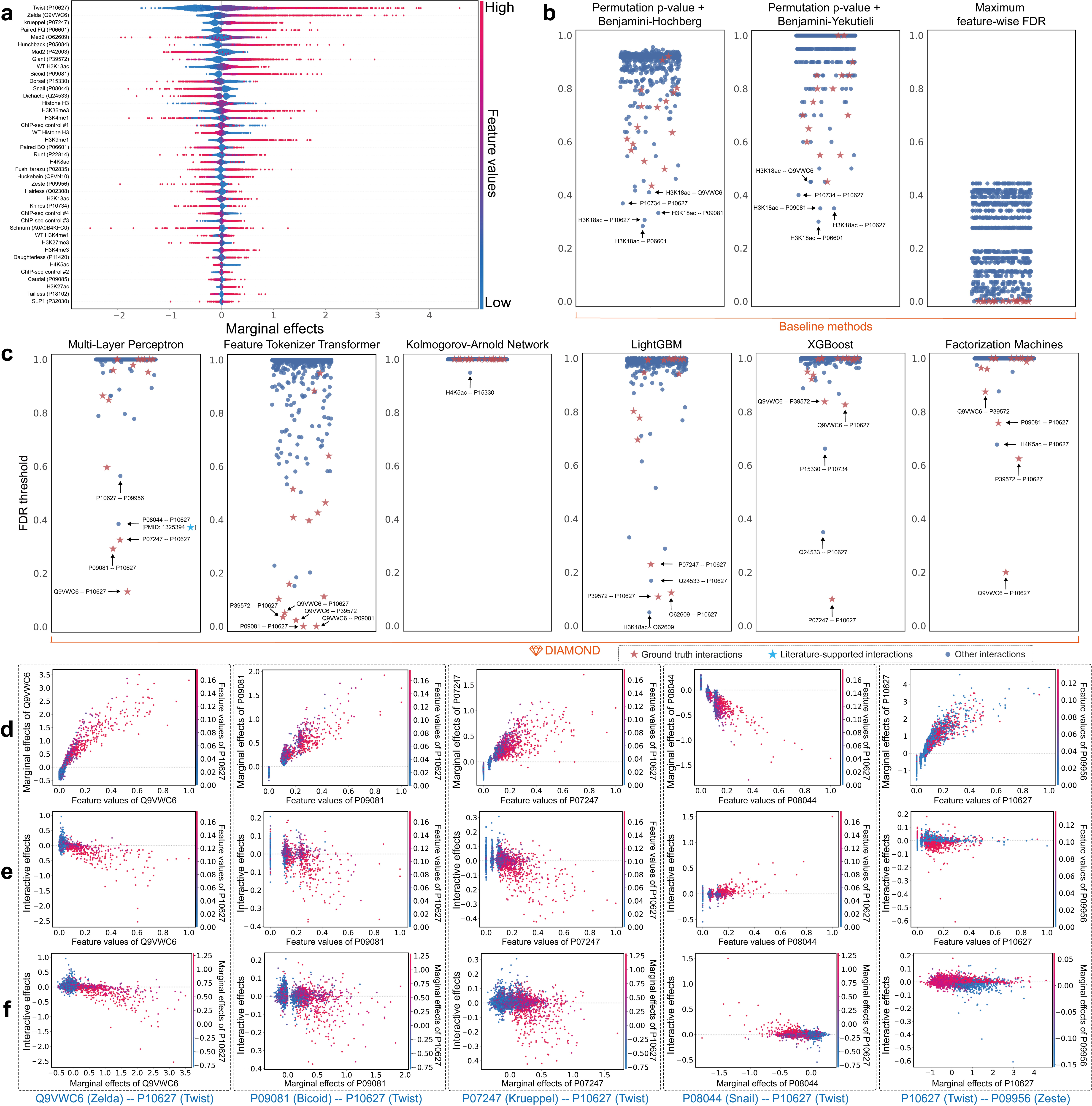}
\caption{
{\bf Evaluating \methodname\ on a real \textit{Drosophila} enhancer dataset.}
(\textbf{a}) Each feature contributes differently to predicting enhancer status, as measured by the Expected Gradient scores in the MLP model.
(\textbf{b}) \methodname\ is compared against three baseline methods.
The annotated transcription factors are labeled by their UniProt identifiers.
The red stars indicate well-characterized physical interactions in early \textit{Drosophila} embryos as ground truth. 
The blue stars indicate interactions supported by literature evidence, referenced by the accompanying PubMed identifiers.
(\textbf{c}) \methodname\ is used with various ML models to identify important non-additive interactions.
Each possible interaction is measured by the minimum FDR threshold cutoff at which it is selected, with the top five interaction annotated.
The top interactions reported by the MLP model are qualitatively evaluated from three aspects: the contribution of feature values to the (\textbf{d}) marginal and (\textbf{e}) interaction importance measure, and (\textbf{f}) the contribution of the marginal importance measures to the interaction importance measure.
\label{fig:enhancer}}
\end{figure}
We then evaluated the performance of \methodname\ on a \textit{Drosophila} enhancer dataset \citep{basu:iterative} to investigate the relationship between enhancer activity and DNA occupancy for transcription factor (TF) binding and histone modifications in \textit{Drosophila} embryos.
We used a quantitative study of DNA occupancy for $23$ TFs and $13$ histone modifications with the enhancer status labeled for $7,809$ genomic sequence samples from blastoderm \textit{Drosophila} embryos (Fig. \ref{fig:diabetes}\textbf{a}).
The enhancer status for each sequence is binarized as the response, depending on whether the sequence drives patterned expression in blastoderm embryos. To predict the enhancer status, the maximum value of normalized fold-enrichment \citep{li:transcription} of ChIP-seq and ChIP-chip assays for each TF or histone modification served as our features.
We considered different types of ML models: MLP, FT-Transformer, KAN, LightGBM, XGBoost, and FM.
We assessed the trained ML model using 5-fold cross-validation and selected the one with the best performance for model interpretation and interaction discovery.
For robustness, we repeated the experiment 20 times using different random seeds. 
Each repetition involved knockoff generation, ML model training, and interaction-wise FDR estimation.

We started by comparing the identified interactions against a list of well-characterized physical interactions in the early \textit{Drosophila} embryos, collected by \cite{basu:iterative}.
These interactions have been identified over decades and have been confirmed to play a critical role in regulating spatial and temporal patterning \citep{rivera:gradients}.
If \methodname\ is successful in identifying important interactions while controlling the FDR, then the identified interactions should considerably overlap with these previously reported physical interactions.
Meanwhile, it is important to acknowledge that this list of interactions, proposed years ago, represents only a partial set of the true interactions. 
Therefore, a detected interaction not appearing in the list does not necessarily indicate it is incorrect.
Our results show that three of the top five interactions reported by \methodname\ are included in the ground truth list for MLP and tree-based models, four out of five for FM, and all five for the FT-Transformer (Fig. \ref{fig:enhancer}\textbf{c}).
The only exception is KAN, which reported all but one interactions at the same FDR levels.
In comparison, the baseline methods exhibit significantly different behaviors (Fig. \ref{fig:enhancer}\textbf{b}). 
Two baseline methods, which are based on permutation-based interaction-wise p-values combined with the Benjamini-Hochberg and the Benjamini-Yekutieli procedures respectively, reported no ground truth interactions among its top five identifications. 
Another method, which aggregates feature-wise FDR, reported ground truth interactions along with a large number of non-ground truth interactions.

Next, we investigated the identified interactions that are not included in the ground truth list.
For example, we investigated the interaction between the TFs Snail (UniProt ID: P08044) and Twist (UniProt ID: P10627) identified by \methodname+MLP through databases 
containing assorted experimentally verified interactions \citep{liska:tflink} and literature \citep{ip:dorsal}.
Research indicates that Snail represses Twist targets because their target sequences are very similar, and their binding is mutually exclusive.
This literature evidence is further supported by a qualitative evaluation (Fig. \ref{fig:enhancer}\textbf{d-f}, Fig. \ref{fig:supp:enhancer_transformer}, Fig. \ref{fig:supp:enhancer_lightgbm}, and Fig. \ref{fig:supp:enhancer_xgboost}).
Specifically, we examined the Snail--Twist interaction in terms of the contribution of feature values to the enhancer status prediction, observing that high Twist expression suppresses the contribution of Snail to enhancer activation.

Furthermore, we investigated the remaining identifications that lack supporting ground truth or literature evidence. 
We discovered that these interactions can be logically explained through transitive effects. 
Specifically, even without a direct interaction between TF1 and TF3, we might still expect a strong interaction between them if solid interactions exist between TF1 and TF2 and between TF2 and TF3.
On this basis, for example, the interaction between the TFs Twist (UniProt ID: P10627) and Zeste (UniProt ID: P09956), identified by \methodname+MLP, can be classified as a transitive interaction, supported by experimentally validated interactions \citep{yevshin:gtrd}: 
(1) the interaction between Twist and Ultrabithorax (UniProt ID: P83949), Decapentaplegic (UniProt ID: P07713), Pho (UniProt ID: Q8ST83), Raf (UniProt ID: P11346), Psc (UniProt ID: P35820), Trr (UniProt ID: Q8IRW8)
and (2) the interaction between these proteins and Zeste.
Analogously, the interaction between the TFs Twist (UniProt ID: P10627) and Dichaete (UniProt ID: Q24533), identified by \methodname\ together with two tree-based models, can also be supported by transitive interactions \citep{yevshin:gtrd}: 
(1) the interaction between Twist and Pho, Raf, Psc, Trr, Myc (UniProt ID: Q9W4S7), Trx (UniProt ID: P20659 )
and (2) the interaction between these proteins and Dichaete.

Finally, \methodname\ is able to identify non-additive interactions with various interactive patterns (Fig. \ref{fig:enhancer}\textbf{d-f}). 
Among the top five interactions reported by \methodname\ with MLP, the Zelda--Twist interaction, the Bicoid--Twist interaction, and the Krueppel--Twist interaction exhibit synergistic effects. 
The high expression of both TFs in these interactions reinforces the contribution of each individual factor to enhancer activation.
In contrast, the Snail--Twist interaction, which was mentioned earlier, exhibits repressive effects, where high expression of one TF suppresses the contribution of the other to enhancer activation.
Lastly, the Twist--Zeste interaction exhibits alternative effects, where high expression of either TF enhances the contribution of each to enhancer activation.

\subsection{\methodname\ understands mortality risk factors in health outcomes}
\label{sec:result:mortality}
\begin{figure}[H]
\centering
\includegraphics[width=1.0\textwidth]{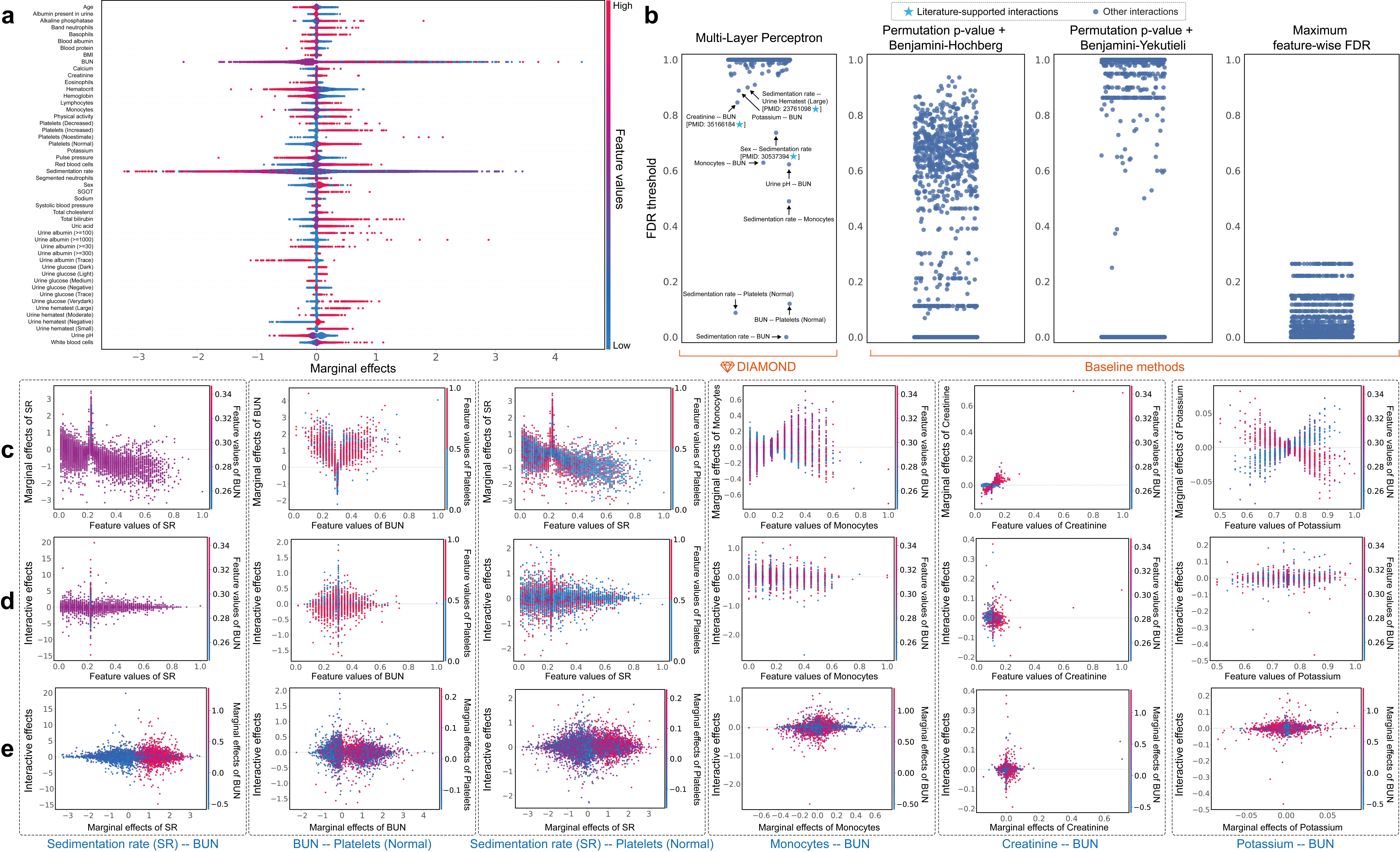}
\caption{
{\bf Evaluating \methodname\ on a real mortality risk dataset.}
(\textbf{a}) Each feature contributes differently to predicting the mortality status, as measured by the Expected Gradient scores.
It is worth mentioning that two marginally important features do not necessarily result in important interactions, as anticipated in \methodname's design.
(\textbf{b}) \methodname\ is used with the MLP model to identify important non-additive interactions and is compared against other baseline methods.
Each possible interaction is measured by the minimum FDR threshold cutoff at which it is selected, with the top five interaction annotated.
The blue stars indicate interactions supported by literature evidence, referenced by the accompanying PubMed identifiers.
The survival loss function for predicting mortality restricts the modeling to DNNs, and we use MLP specifically due to memory limitations with other DNN models.
The top interactions reported by \methodname\ are qualitatively evaluated from three aspects: the contribution of feature values to the (\textbf{c}) marginal and (\textbf{d}) interaction importance measure, and (\textbf{e}) the contribution of the marginal importance measures to the interaction importance measure.
\label{fig:mortality}}
\end{figure}
We lastly evaluated the performance of \methodname\ on a real mortality risk dataset \citep{cox:plan} to investigate the relationship between mortality risk factors and long-term health outcomes within the US population. 
We used a mortality dataset from the National Health and Nutrition Examination Survey (NHANES I) and NHANES I Epidemiologic Follow-up Study (NHEFS).
The dataset incorporates $35$ clinical and laboratory measurements from $14,407$ US participants between 1971 and 1974 (Fig. \ref{fig:mortality}\textbf{a}).
Note that some of the features are presented categorically, which makes model's trained from this data challenging to interpret. 
We addressed this problem by converting each categorical feature into a list of binary features using one-hot encoding.
The dataset also reports the mortality status of participants as of 1992, with $4,785$ individuals known to have passed away before 1992.
The task is to construct an MLP-based model to predict mortality status using a survival loss function.
We assessed the trained model using 5-fold cross-validation and selected the one with the best performance for model interpretation and interaction discovery.
For robustness, we repeated the experiment 20 times using different random seeds. 
Each repetition involved knockoff generation, ML model training, and interaction-wise FDR estimation.

Given the absence of ground truth interactions for this task, we began by evaluating the interactions identified by \methodname\ through literature support.
Our analysis shows that three of the top ten selected interactions are directly supported by existing literature (Fig. \ref{fig:mortality}\textbf{b}).
For example, we investigated the interaction between sex and sedimentation rate (SR) and found that, although a high SR is strongly associated with a higher risk of overall mortality, this association can be attenuated by sex \citep{fest:erythrocyte}.
Additionally, we investigated the interaction between creatinine and the blood urea nitrogen (BUN) identified by \methodname.
Research indicates that the BUN/creatinine ratio is recognized to have a nonlinear association with all-cause mortality and a linear association with cancer mortality \citep{shen:blood}.
The literature evidence is further supported by qualitative evaluation (Fig. \ref{fig:mortality}\textbf{c-e}).
Specifically, we examined the BUN-creatinine interaction in terms of the contribution of feature values to mortality status prediction, observing that when the creatinine level exceeds a certain level, a high BUN level further increases the mortality risk. 
In contrast, the baseline methods display significantly different behavior, reporting an overwhelming number of interactions at the significant FDR level, which appears to overestimate false positives.

Finally, we probed into the remaining identifications that lack supporting literature evidence. 
As in the \textit{Drosophila} analysis, we discovered that these interactions can be logically explained through transitive effects. 
For example, the interaction between BUN and potassium can be justified by combining the following two established facts:
(1) BUN level is indicative of chronic kidney disease development \citep{collins:association}, and
(2) Patients with chronic kidney diseases show a progressively increasing mortality rate with abnormal potassium levels \citep{seki:blood}.
Additionally,  the interaction between BUN and SR reflects the fact that 
(1) The creatinine level and SR are strongly associated to clinical pathology types and prognosis of patients \citep{liang:serum}, and 
(2) Creatinine and BUN have a nonlinear association \citep{shen:blood}.

\section{Discussion}
\label{sec:discussion}
In this study, we aim to enable rigorous data-driven scientific discoveries, which are crucial in the present of massive data sets.
For this purpose, we propose \methodname, an error-controlled interaction discovery method designed to work with a variety of ML models.
The key novelties of \methodname\ are threefold. 
First, \methodname\ achieves FDR control using carefully designed knockoffs, without relying on p-values, which are often difficult to obtain in generic ML models.
Second, \methodname\ distills non-additive interaction effects, thereby maintaining FDR control at the target level.
This non-additivity distillation step is critical because naive application of off-the-shelf feature interaction importance measures fail to control the FDR in this setting.
Third, \methodname\ is versatile in discovering important non-additive interactions across a wide range of ML models, different knockoff designs, and various interaction importance measures, all while guaranteeing FDR control.
We have applied \methodname\ to various simulated and real datasets to demonstrate its empirical utility, proving it to be the only valid solution for error-controlled data-driven scientific discovery compared to other alternative methods.

Methodologically, our approach provides a path toward enhancing transparency in complex ML models.
The complexity of ML models has long posed a tradeoff between predictability and interpretability.
Specifically, sophisticated ML models such as DNNs excel at detecting subtle relationships and patterns within complex data. 
However, their \quotes{black-box} nature poses challenges in mission-critical domains where error tolerance is minimal.
By providing FDR control in this setting, we can better understand why and how models make predictions, thereby enabling more effective use of ML models to gain human-understandable insights.
\methodname\ can help diagnose potential biases in ML models, contributing positively to the integrity, accountability, and fairness required when employing AI technologies.

Lastly, this study highlights several promising directions for future research.
First, we observe that \methodname\ tends to be conservative. 
This suggests a future direction for improvement in enhancing statistical power through further refinement of the FDR estimation process.
Previous efforts, like the knockoff-tailored plugin pairwise-coupling layer in MLP models, have shown superior performance in statistical power \cite{lu:deeppink}.
However, these previously described methods are specific to certain types of ML models and may not be generally applicable to all generic ML models, indicating the need for improvement to broaden their applicability.

Secondly, we observe that \methodname\ cannot distinguish between direct interactions and interactions caused by transitive effects. 
In reality, the former is scientifically more interesting than the latter, though both fit the definition of non-additive interactions.
Specifically, even without a direct interaction between two features, we might still expect a strong interaction between them if solid interactions exist individually between each of these features and a third feature.
Though transitive interactions are non-additive, for the purpose of scientific discovery, a potential research direction is to identify FDR-controlled direct interactions exclusively, using a causal inference framework \citep{luo:causal}.
Thirdly, we observe that different ML models exhibit different levels of performance in identifying important interactions.
Briefly, models with stronger modeling capabilities, such as transformer models, are better at identifying important interactions compared to models with weaker capabilities, such as FMs and CNNs.
As independent studies have highlighted, the sensitivity of interaction detection depends on the importance measures of the interactions, which are critically influenced by the quality of the model \citep{adebayo:sanity}.
A potential research direction is to automatically determine the best ML models for interaction detection, rather than focusing solely on modeling accuracy.

Finally, this study primarily focuses on discovering pairwise interactions.
While detecting pairwise interactions holds practical significance in various biological contexts, the discovery of higher-order interactions can naturally provide deeper insights for explaining genetic mechanisms, diseases, and drug effects in healthcare domains.
This problem is highly challenging in practice due to the exponentially large search space. 
A potential research direction for future studies is to effectively reduce the search space to generalize \methodname\ for practical discovery of higher-order interactions.

In conclusion, \methodname\ enables error-controlled detection of non-additive feature interactions in the context of a variety of ML models.
The versatility and flexibility of \methodname\ make it widely applicable in high-stakes and error-intolerant domains where interpretability and statistical rigor are needed.
The robustness against misspecified ML models and low-quality knockoffs ensures the reliability of \methodname, even when erroneously applied in extreme cases.
We believe that this powerful tool will pave the way for the broader deployment of machine learning models in scientific discovery and hypothesis validation.

\section{Methods}
\label{sec:methods}

\subsection{FDR control with the knockoffs}
\label{sec:methods:knockoffs}

\methodname\ achieves FDR control by leveraging the model-X knockoffs framework \citep{barber:controlling, candes:panning}, which was proposed in the setting of error-controlled feature selection.
The core idea of knockoffs is to generate dummy features that perfectly mimic the empirical dependence structure among the original features but are conditionally independent of the response given the original features.
Briefly speaking, the knockoff filter achieves FDR control in two steps: (1) construction of knockoff features and (2) filtering using knockoff statistics.

For the first step, the knockoff features are defined as follows: 
\begin{definition}[Model-X knockoff \citep{candes:panning}]
\label{def:knockoff}
The model-X knockoff features for the family of random features $\mathbf{X}=(X_{1},\dots,X_p)$ are a new family of random features $\mathbf{\tilde{X}}=(\tilde{X}_{1},\dots,\tilde{X}_{p})$ that satisfy two properties:
        \begin{enumerate}
        \item $(\mathbf{X},\tilde{\mathbf{X}})_{\text{swap}(\mathcal{S})}\stackrel{d}{\text{=}}(\mathbf{X},\tilde{\mathbf{X}})$ for any subset $\mathcal{S}\subset \left \{ 1,\dots, p \right \}$, where $\text{swap}(\mathcal{S})$ means swapping $X_{j}$ and $\tilde{X}_{j}$ for each $j \in \mathcal{S}$ and $\stackrel{d}{\text{=}}$ denotes equal in distribution, and
        \item $\tilde{\mathbf{X}} \dperp \mathbf{Y}|\mathbf{X}$, i.e., $\tilde{\mathbf{X}}$ is independent of response $\mathbf{Y}$ given feature $\mathbf{X}$. \\
          \end{enumerate}
\end{definition} 

According to Definition~\ref{def:knockoff}, the construction of the knockoffs must be independent of the response $\mathbf{Y}$.
Thus, if we can construct a set $\tilde{X}$ of model-X knockoff features properly, then by comparing the original features with these control features, FDR can be controlled at target level $q$.
In the Gaussian setting, \ie $\mathbf{X}\sim \mathcal{N}(0,\mathbf{\Sigma} )$ with covariance matrix $\mathbf{\Sigma} \in \mathbb{R}^{p \times p}$, the model-X knockoff features can be constructed easily:
\begin{equation}
\begin{aligned}
\tilde{\mathbf{X}}|\mathbf{X} \sim N\big(\mathbf{X} - &\mathrm{diag}\{\mathbf{s}\} \mathbf{\Sigma}^{-1} \mathbf{X}, 2\mathrm{diag}\{\mathbf{s}\} - \mathrm{diag}\{\mathbf{s}\} \mathbf{\Sigma}^{-1} \mathrm{diag}\{\mathbf{s}\}\big)
\end{aligned}
\label{eq:knockoff1}
\end{equation}
where $\mathrm{diag}\{\mathbf{s}\}$ is a diagonal matrix with all components of $\mathbf{s}$ being positive such that the conditional covariance matrix in Equation~\ref{eq:knockoff1} is positive definite.
As a result, the original features and the model-X knockoff features constructed by Equation~\ref{eq:knockoff1} have the following joint distribution:
\begin{equation}
\label{eq:knockoff2}
(\mathbf{X},\tilde{\mathbf{X}})\sim \mathcal N\left ( \begin{pmatrix}
\mathbf{0}\\
\mathbf{0}
\end{pmatrix},\begin{pmatrix}
\mathbf{\Sigma} & \mathbf{\Sigma}-\mathrm{diag}\{\mathbf{s}\}\\
\mathbf{\Sigma}-\mathrm{diag}\{\mathbf{s}\} & \mathbf{\Sigma}
\end{pmatrix} \right )
\end{equation}

With the constructed knockoff $\tilde{\mathbf{X}}$, feature importances are quantified by computing the knockoff statistics $W_{j}=g_{j}(Z_{j},\tilde{Z}_{j})$ for $1\leq j\leq p$, where $Z_{j}$ and $\tilde{Z}_{j}$ represent feature importance measures for the $j$-th feature $X_{j}$ and its knockoff counterpart $\tilde{X}_{j}$, respectively, and $g_{j}(\cdot,\cdot)$  is an antisymmetric function satisfying $g_{j}(Z_{j},\tilde{Z}_{j})=-g_{j}(\tilde{Z}_{j},Z_{j})$.
The knockoff statistics $W_{j}$ should satisfy a coin-flip property such that swapping an arbitrary pair $X_{j}$ and its knockoff counterpart $\widetilde{X}_{j}$ only changes the sign of $W_{j}$ but keeps the signs of other $W_{k}$ ($k \neq j$) unchanged \citep{candes:panning}.
A desirable property for knockoff statistics $W_j$'s is that important features are expected to have large absolute values, whereas unimportant ones should have small symmetric values around 0. 

Finally, the absolute values of the knockoff statistics $|W_{j}|$'s are sorted in decreasing order, and FDR-controlled features are selected whose $W_{j}$'s exceed some threshold $T$.
In particular, the choice of threshold $T$ follows $T=\min\left \{ t\in \mathcal{W},\frac{1+|\left \{ j:W_j \leq -t \right \}|}{|\left \{ j: W_j\geq t \right \}|}\leq q \right \}$ where $\mathcal{W}=\left \{ |W_{j}|:1\leq j\leq p \right \} \char`\\ \left \{ 0\right \}$ is the set of unique nonzero values from $|W_{j}|$'s and $q\in (0,1)$ is the desired FDR level specified by the user. 

\subsection{Measuring non-additive interaction effect}
\label{sec:methods:distillation}
As a key precursor to FDR estimation, \methodname\ quantifies feature interactions from trained ML models and produces a ranked list of these interactions, with higher-ranked interactions indicating greater importance.
For notational simplicity, we use indices for both original features and knockoffs as $\left \{ 1,2,\cdots,2p \right \}$, 
with $\left \{ 1,\cdots,p \right \}$ and $\left \{ p+1,\cdots,2p \right \}$ corresponding to the original features
and their respective knockoff counterparts.
Here, we define $\mathbf{E}^{\text{2D}}=\left [ e_{ij} \right ]_{i,j=1}^{2p}\in \mathbb{R}^{2p\times 2p}$ 
as a reported interaction importance measure from existing methods.
There are many feature interaction importance measures available for $\mathbf{E}^{\text{2D}}$, each attributing the prediction influence to feature pairs in different ways.
However, it is important to note that such measures favor pairs where both features are simultaneously important for a model's prediction, rather than capturing the true non-additive interaction effects between the two features \citep{tsang:interpretable}.
Further supported by simulation studies (Fig. \ref{fig:simulation}\textbf{c}), we observed that many feature interaction importance measures tend to assign higher interaction scores to two marginally important but non-interacting features compared to two random ones, even though neither pair has a real interaction, leading to the failure of FDR control.


The direct reason for the interaction importance measure falling short in controlling FDR is its violation of the knockoff filter's assumption. 
Specifically, the knockoff filter requires that the importance scores of knockoff-involving interactions and false interactions have a similar distribution.
To resolve this issue, we introduce a non-additivity distillation procedure to be applied on top of existing interaction importance measures.
Specifically, we consider that a reported interaction importance measure from existing methods comprises a mixture of several factors: label-dependent marginal effects for individual features, label-independent feature biases, independent random noise, and potential non-additive interaction effects between feature pairs.
Thus, the reported interaction between features $i$ and $j$ is represented as:
\begin{equation}
e_{ij} = s_{ij} + g_i(e_i) + g_j(e_j) + b(I_{ij}) + \varepsilon_{ij}
\label{eq:decompose}
\end{equation}
where $\varepsilon_{ij} \in \mathbb{R}$ is random noise independent of both features and predictions, $e_{ij} \in \mathbb{R}$ and $e_i,e_j \in \mathbb{R}$ are reported pairwise and univariate feature importance measures that are dependent on the model's predictions, respectively.
The functions $g_i,g_j: \mathbb{R} \mapsto \mathbb{R}$ adapt univariate feature importance to be compatible with feature interaction importance.
The function $b: \mathbb{R}^{2p} \mapsto \mathbb{R}$ models the feature-specific biases that are independent of the model's predictions, where $I_{ij} \in \left \{ 0,1 \right \}^{2p}$ indicates the presence of feature $i$ and $j$.
Our goal is to identify $s_{ij}$, the potential non-additive interaction effects between features $i$ and $j$.

We formulate the identification of interaction effects $s_{ij}$ as the residuals of a regression task: 
\begin{equation}
\min_{b,g_1,g_2,\cdots} \sum_{i<j} w_{ij} \cdot \left \| e_{ij} - g_i(e_i) - g_j(e_j) - b(I_{ij}) \right \|^2
\label{eq:regression}
\end{equation}
where $w_{ij}>0$ is the conditional probability of being either original-only (\ie $i,j\leq p$) or knockoff-involving (\ie $i> p$ or $j> p$) feature pairs given two univariate feature importance measures $e_i$ and $e_j$, estimated by a logistic regression model \citep{freedman:weighting}.
The rationale is based on the important observation that most feature pairs do not exhibit non-additive interactions, especially those involving knockoff features. 
Therefore, we want to focus more on potential non-additive interactions that have large univariate feature importance, as important interactions naturally consist of significant marginal features.
In this study, we parameterize the functions $b: \mathbb{R}^{2p} \mapsto \mathbb{R}$ and $g_i,g_j: \mathbb{R} \mapsto \mathbb{R}$ using generalized additive models and optimize Eq. \ref{eq:regression} using the pyGAM library \citep{serven:pygam}.

\subsection{FDR control for interactions}
\label{sec:methods:fdr}

After calculating the non-additive interaction effects using Eq. \ref{eq:regression}, we denote the resultant set of interaction effects as $\Gamma =\left \{s_{ij} |i<j, i\neq j-p \right \}$.
We arrange $\Gamma$ in decreasing order and select interactions for which the interaction effect, $\Gamma_{j}$, exceeds some threshold, $T$. 
This selection ensures that the chosen interactions adhere to a desired FDR level $q\in (0,1)$.

However, a point of complexity is introduced due to the heterogeneous interactions, which include original-only and knockoff-involving interactions. 
The latter further comprises original-knockoff, knockoff-original, and knockoff-knockoff interactions.
Following the strategy outlined by \cite{walzthoeni:false}, the threshold $T$ is determined by:
\begin{equation}
  \begin{aligned}
    &T=\min\left \{ \vphantom{\int_1^2} t\in \mathcal{T}, \right. 
    \left. \frac{|\left \{ j:\Gamma_j \geq t, j \in \mathcal{K} \right \}|-2\cdot |\left 
    \{ j:\Gamma_j \geq t, j \in \mathcal{KK} \right \}|}{|\left \{ j: \Gamma_j\geq t, j \notin \mathcal{K} \text{ and } j \notin \mathcal{KK} \right \}|}\leq q \right \}
  \end{aligned}
  \label{eq:interaction_fdr}
\end{equation}
where $\mathcal{K}$ and $\mathcal{KK}$ respectively denote the sets of interactions that include at least one 
knockoff feature and both knockoff features, while $\mathcal{T}$ refers to the set of unique nonzero values present in $\Gamma$.
Refer to Sec. \ref{sec:supp:fdr} for more detailed description of FDR estimation.

\subsection{ML models}
\label{sec:methods:models}
\methodname\ is designed to be compatible with a wide range of ML models. 
To demonstrate the broad applicability of \methodname, we use it in conjunction with machine learning models from methodologically distinct categories, including DNN models, Kolmogorov-Arnold networks (KANs) \citep{liu:kan}, tree-based models, and factorization-based models.

For DNNs, we selected three representative models: multi-layer perceptron (MLP), convolutional neural network (CNN), and transformer models.
It is worth noting that CNNs are not well-suited for modeling tabular data and are included here as an example of model misspecification.
We configured a pyramid-shaped MLP model with the exponential linear unit activation and four hidden layers, each having neuron sizes of 2p, p, p/2, and p/4 respectively, where p denotes the input dimensionality.
We configured a CNN model consisting of two convolutional layers. The first layer has one input channel and 32 output channels, while the second layer has 32 input channels and 64 output channels, followed by max pooling. Batch normalization and ReLU activation are applied after each convolution. The output is then flattened and passed through two fully connected layers, mapping to 128 units before reaching the output dimension.
For transformer models, we used FT-Transformer \citep{gorishniy:revisiting}, a model specifically designed for tabular data. By default, it consists of 6 layers, 8 attention heads, a 32-dimensional feedforward network, and a 0.1 dropout rate.
All DNN models follow the guidance of \cite{lu:deeppink} and include a plugin pairwise-coupling layer, connecting each original feature with its knockoff counterpart in a pairwise fashion to maximize statistical power (See details in Sec. \ref{sec:supp:mlp}).

For KANs, we configured a two-layer model, with a hidden layer of width p/2, followed by a mapping to the output dimension.
For tree-based models, we employed two widely used gradient-boosted decision tree representatives---XGBoost \citep{chen:xgboost} and LightGBM \citep{ke:lightgbm}.
We used the implementations provided by the two libraries with default settings.
For factorization-based models, we used a widely adopted representative method---Factorization machines (FM) \citep{rendle:factorization}.
We used 2-Way FM implemented in the xLearn library \citep{ma:xlearn} with default settings, which models up to second-order feature interactions.

\subsection{Model interpretation}
\label{sec:methods:interpretation}

\methodname\ utilizes representative importance measures to interpret the trained ML models and elucidate the relationship between features and responses.
For DNN and KAN models, we employed a representative model-agnostic interaction importance measure, Expected Hessian \citep{erion:improving}, to clarify the relationships between features and responses without making assumptions about any specific model architecture (See details in Sec. \ref{sec:supp:model-agnostic}).
To demonstrate \methodname's flexibility, we also used Integrated Hessian \citep{janizek:explaining}, another state-of-the-art, model-agnostic interaction importance measure, alongside a model-specific interaction importance measure derived from the model weights, specifically designed for the MLP architecture (See details in Sec. \ref{sec:supp:model-specific}).
Both model-agnostic interaction importance measures, Expected Hessian and Integrated Hessian, are computed using the Path-Explain library \citep{janizek:explaining}. 

For tree-based models such as XGBoost and LightGBM, we used TreeSHAP \citep{lundberg:local}, a Shapley value-based interpretation method specifically designed for tree-based ML models, to clarify the relationships between features and responses.
We used the implementation provided by the SHAP library \citep{lundberg:unified} to calculate both univariate and pairwise feature importance.
For factorization-based models, since 2-Way FM is explicitly modeled as the weighted sum of univariate and pairwise feature interactions, we used the learned coefficients of these interactions as the corresponding importance measures.

\subsection{Knockoff generation}
\methodname\ relies on knockoffs to control the FDR.
It is worth mentioning that conventional knockoffs are limited to Gaussian settings, which may not be applicable in many practical scenarios.
In this study, we focused on the state-of-the-art knockoff design without assuming any specific feature distribution.
In particular, we considered commonly-used non-Gaussian knockoff generation methods such as KnockoffsDiagnostics \citep{blain:when}, KnockoffGAN \citep{jordon:knockoffgan}, Deep knockoffs \citep{romano:deep}, and VAE knockoffs \citep{liu:auto-encoding}.
Furthermore, to assess \methodname's robustness in the presence of poorly generated knockoffs, we created invalid knockoffs by independently permuting each feature across samples, thus violating the definition of knockoffs outlined in Sec. \ref{sec:methods:knockoffs}.

\subsection{Alternative FDR estimation methods}
\label{sec:methods:baselines}

We evaluate the performance of \methodname\ in comparison to three baseline methods. 
For the first two baseline methods, we employ a permutation-based approach to calculate the interaction-wise FDR. 
Specifically, this involves using a previously described permutation procedure tailored for neural networks to assess the significance of interactions and calculate permutation p-values \citep{cui:gene}.
The permutation p-values are combined with the Benjamini–Hochberg procedure \citep{benjamini:controlling} or the Benjamini-Yekutieli procedure \citep{benjamini:control} to estimate the FDR.
The difference between the Benjamini–Hochberg procedure and the Benjamini-Yekutieli procedure is that the former assumes the p-values are independent or positively dependent, while the latter imposes no assumptions on the dependencies among p-values.

For the second baseline method, we consider an ensemble-based approach that represents interaction-wise FDR as the aggregation of feature-wise FDR.
This approach follows the intuition that an important feature interaction is composed of important univariate features.
Specifically, we use a previously described knockoff-based procedure tailored for neural networks to estimate the feature-wise FDR of each univariate feature \citep{lu:deeppink}. 
We then approximate the interaction-wise FDR as the maximum of the two comprising univariate feature-wise FDRs.

\subsection{Code and data availability}
\label{sec:availability}

The software implementation and all models described in this study are available at \url{https://github.com/batmen-lab/diamond}.
All public datasets used for evaluating the models are available at \url{https://github.com/batmen-lab/diamond/data}.

%
%
%
%
\section*{Declarations}


\begin{itemize}
\item The research is supported by the Canadian NSERC Discovery Grant RGPIN-03270-2023.
\item The authors declare that they have no conflict of interest.
\item There is no ethics approval and consent to participate involved in this study.
\item There is no consent for publication involved in this study.
\item Author contribution: W.C. and Y.J. implemented the code and did the analysis. W.C. and Y.J. set up and preprocessed the datasets. W.C., Y.J. and Y.L. prepared the figures. Y.L. wrote the manuscript. All authors participated the discussion. All authors reviewed the manuscript.
\end{itemize}

\bibliography{refs.bib} 

\newpage
\appendix
\section{Appendix}
\label{sec:supp}
\definecolor{llgray}{RGB}{240,240,240}

\setcounter{figure}{0} \renewcommand{\thefigure}{A.\arabic{figure}}
\setcounter{table}{0} \renewcommand{\thetable}{A.\arabic{table}}
\setcounter{section}{0} \renewcommand{\thesection}{A.\arabic{section}}

\section{Knockoff-tailored multi-layer perceptron model}
\label{sec:supp:mlp}
\begin{figure}[H]
\centering
\captionsetup{justification=centering}
\includegraphics[width=0.7\textwidth]{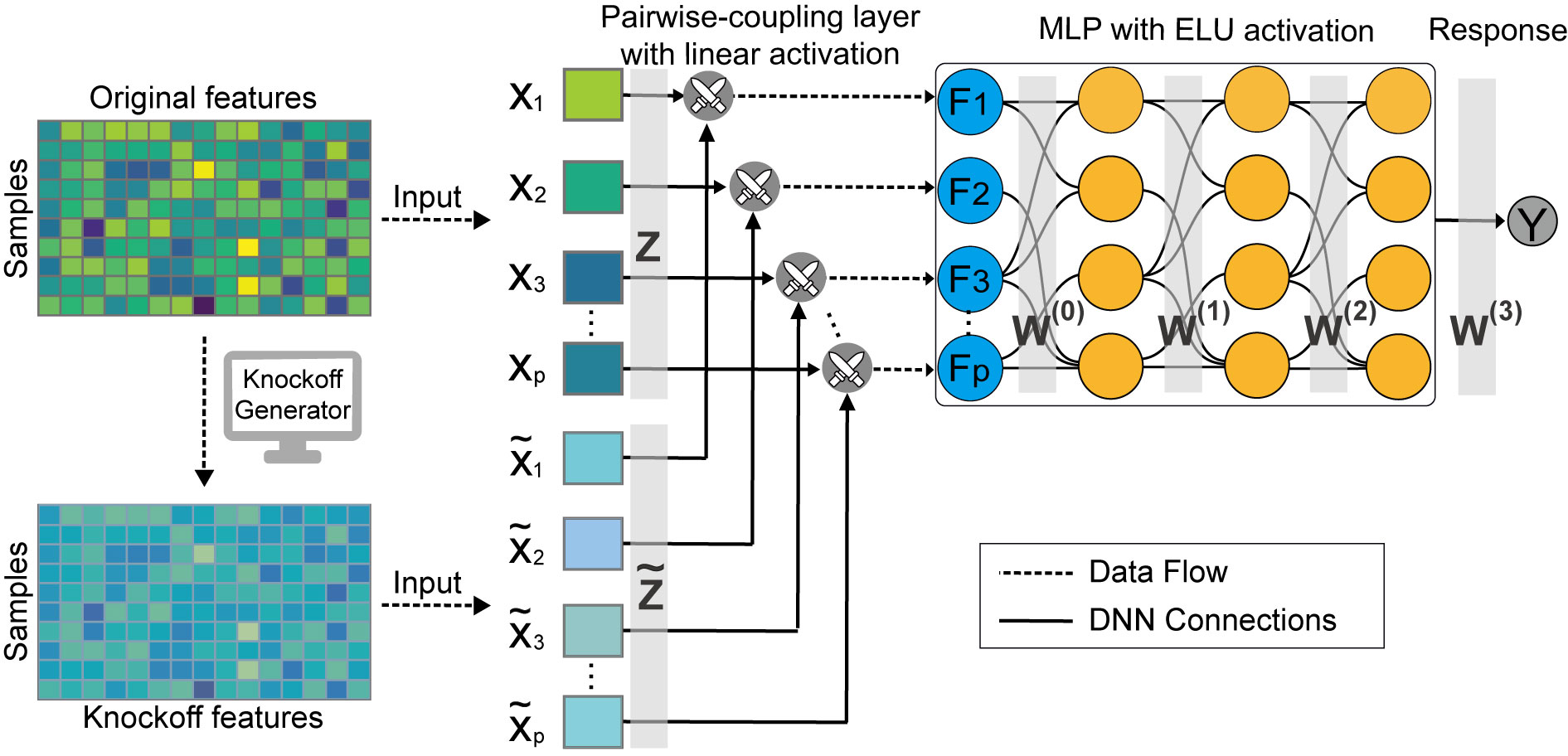}
\caption{
{\bf Illustration of knockoff-tailored MLP model.}
\label{fig:supp:mlp}}
\end{figure}
By following the guidance of \cite{lu:deeppink}, \methodname\ utilizes a knockoff-tailored multi-layer perceptron (MLP) model that includes a plugin pairwise-coupling layer, connecting each original feature with its knockoff counterpart in a pairwise fashion to maximize statistical power.
Specifically, the model takes as input an augmented data matrix $(\mathbf{X}, \tilde{\mathbf{X}}) \in \mathbb{R}^{n \times 2p}$, created by concatenating the original data matrix $\mathbf{X} \in \mathbb{R}^{n \times p}$ with its knockoffs $\tilde{\mathbf{X}} \in \mathbb{R}^{n \times p}$.
The augmented data matrix is fed to an off-the-shelf MLP model through a plugin pairwise-coupling layer composed of $p$ filters, encapsulated by $\mathbf{F}=(F_{1},\cdots, F_{p}) \in \mathbb{R}^{p}$, where each $j$-th filter connects feature $X_{j}$ and its knockoff counterpart ${\tilde{X}}_{j}$, as shown in Fig. \ref{fig:supp:mlp}.

The filter weights, $\mathbf{Z}\in \mathbb{R}^{p}$ and $\mathbf{\tilde{Z}}\in \mathbb{R}^{p}$ are initialized 
identically and engage in a competitive dynamic via pairwise connections during the DNN training.
Additionally, we employ a linear activation function in the pairwise-coupling layer to stimulate competition between different features. 
The outputs of the filters are subsequently channeled into an MLP model that learns to map to the response $\mathbf{Y}$.
In this study, we chose an MLP architecture with the exponential linear unit (ELU) activation function and four hidden layers.
Letting $L$ be the number of hidden layers and $p_l$ denote the number of neurons in the $l$-th layer of the MLP 
--- where $p_0=p$ --- we accordingly define the weight matrices of the input layer, hidden layers, and the output layer 
in the MLP as $\mathbf{W}^{(0)}\in \mathbb{R}^{p \times p_1}$, $\mathbf{W}^{(l)}\in \mathbb{R}^{p_l \times p_{l+1}}$,
$\mathbf{W}^{(L)}\in \mathbb{R}^{p_L \times 1}$, respectively.
With these notations, the response $\mathbf{Y}$ is expressed as follows:
\begin{equation}
\begin{aligned}
& \mathbf{h}^{(0)} = \mathbf{F}, \\
& \mathbf{h}^{(l)}=\text{ELU}\left ( \mathbf{W}^{(l-1)} \mathbf{h}^{(l-1)}+\mathbf{b}^{(l-1)} \right ), \text{ for } l=1,\cdots,L \\
& \mathbf{Y} = \mathbf{W}^{(L)} \mathbf{h}^{(L)}+\mathbf{b}^{(L)}
\end{aligned}
\label{eq:mlp}
\end{equation}
where $\text{ELU}(\cdot)$ refers the ELU function, and $\mathbf{b}^{(l)}\in \mathbb{R}^{p_l }$ signifies the bias vector in the $l$-th layer.
While we use this specific model, \methodname's overall process is versatile and fully applicable to any off-the-shelf DNN architecture.

\section{Model-specific interaction importance measure}
\label{sec:supp:model-specific}
The model-specific interaction importance measure is based on the knockoff-tailored MLP model design and notation described in Sec. \ref{sec:supp:mlp}, 
Given the pairwise-coupling layer weights: $\mathbf{Z} \in \mathbb{R}^{p}$ and $\tilde{\mathbf{Z}} \in \mathbb{R}^{p}$ as well as the MLP weights in the $l$-th layer: $\mathbf{W}^{(l)}$, the model-specific interaction importance measure decomposes into two factors:
(1) the relative importance between the original feature and its knockoff counterpart, encoded by concatenated 
filter weights $\mathbf{Z}^{\text{Agg}}=(\mathbf{Z}, \tilde{\mathbf{Z}}) \in \mathbb{R}^{2p}$, and
(2) the relative importance among all $p$ features, encoded by the weight matrix 
$\mathbf{W}^{(0)}\in \mathbb{R}^{p \times p_1}$ and the aggregated weights 
$\mathbf{W}^{\text{Agg}}=\prod_{i=1}^{L} \mathbf{W}^{(i)}\in \mathbb{R}^{p_1}$. 
(See \cite{garson:interpreting} for theoretical insights regarding $\mathbf{W}^{\text{Agg}}$.)

Inspired by \cite{tsang:detecting}, we define the model-specific interaction importance as:
\begin{equation}
e^{\text{2D}}_{ij}=\left ( \mathbf{Z}^{\text{Agg}}_i\mathbf{W}^{\text{INT}}_i\odot \mathbf{Z}^{\text{Agg}}_j\mathbf{W}^{\text{INT}}_j \right )^T \mathbf{W}^{\text{Agg}}
\label{eq:feature:global2d}
\end{equation}
where $\mathbf{W}^{\text{INT}}=({\mathbf{W}^{(0)}}^T, {\mathbf{W}^{(0)}}^T)^T\in \mathbb{R}^{2p \times p_1}$ and 
$\mathbf{W}^{\text{INT}}_j\in \mathbb{R}^{p_1}$ denotes the $j$-th row of $\mathbf{W}^{\text{INT}}$.

Following \cite{lu:deeppink}, we define the model-specific univariate feature importance as:
\begin{equation}
\mathbf{E}^{\text{1D}} = \left [ e^{\text{1D}}_{j} \right ]_{j=1}^{2p} = (\mathbf{Z}\odot\mathbf{W}^{\text{1D}}, \tilde{\mathbf{Z}}\odot\mathbf{W}^{\text{1D}})
\label{eq:feature:global1d}
\end{equation}
where $\mathbf{W}^{\text{1D}}=\mathbf{W}^{(0)}\mathbf{W}^{\text{Agg}} \in \mathbb{R}^{p}$ and $\odot$ denotes entry-wise matrix multiplication.

\section{Model-agnostic interaction importance measure}
\label{sec:supp:model-agnostic}
The model-agnostic interaction importance measure aims to elucidate the relationships between feature pairs and responses without making assumptions about any specific model architecture.
Specifically, we employ two state-of-the-art model-agnostic interaction importance measures---Integrated Hessian \citep{janizek:explaining} and Expected Hessian \citep{erion:improving}.

The Integrated Hessian interaction importance \citep{janizek:explaining} is defined as:
\begin{equation}
e^{\text{2D}}_{ij}=\sum_{x \in \mathbf{X}} \int_{x'}(x_i-x_i')(x_j-x_j')\times
    \int_{\beta=0}^{1} \int_{\alpha=0}^{1} \alpha\beta\nabla^2_{i,j}\mathbf{Y}(x'+\alpha\beta (x-x')) d \alpha d \beta d x'
\label{eq:feature:local2d_integrated}
\end{equation}
where $\nabla^2_{i,j}\mathbf{Y}(x)$ calculates the second derivative of the response $\mathbf{Y}$ with respect to the $i$-th and $j$-th feature of sample $x$ to be explained.
The corresponding univariate feature importance measure, which is compatible with the Integrated Hessian, is Integrated Gradient \citep{sundararajan:axiomatic}:
\begin{equation}
e^{\text{1D}}_{i}=\sum_{x \in \mathbf{X}}\int_{x'}(x_i-x_i')\times \int_{\alpha=0}^{1} \nabla_i\mathbf{Y}(x'+\alpha (x-x')) d \alpha d x'
\label{eq:feature:local1d_integrated}
\end{equation}
where $\nabla_{i}\mathbf{Y}(x)$ calculates the first-order derivative of $\mathbf{Y}$ with respect to the $i$-th feature of input $x$.

The Expected Hessian interaction importance \citep{erion:improving} is defined as:
\begin{equation}
e^{\text{2D}}_{ij}=\sum_{x \in \mathbf{X}} \mathbb{E}_{\alpha\sim U(0,1),\beta\sim U(0,1),x' \in \mathbf{X}} \left[ (x_i-x_i')(x_j-x_j')\alpha\beta \nabla^2_{i,j}\mathbf{Y}(x'+\alpha\beta (x-x')) \right] 
\label{eq:feature:local2d_expected}
\end{equation}
where $U(0,1)$ indicates the uniform distribution. 
The corresponding univariate feature importance measure, which is compatible with the Expected Hessian, is Expected Gradient \citep{erion:improving}:
\begin{equation}
e^{\text{1D}}_{i}=\sum_{x \in \mathbf{X}}\mathbb{E}_{\alpha\sim U(0,1),x' \in \mathbf{X}} \left[ (x_i-x_i')\alpha \nabla_{i}\mathbf{Y}(x'+\alpha (x-x')) \right]
\label{eq:feature:local1d_expected}
\end{equation}

\section{FDR estimation}
\label{sec:supp:fdr}
We estimate the FDR for non-additive interactions by following the strategy proposed by \cite{walzthoeni:false}, which was originally designed to estimate the FDR for cross-linked peptide detection.
Despite the conceptually different application backgrounds, the FDR estimation for cross-linked peptide detection and non-additive interaction detection are essentially analogous. 
Specifically, in the context of cross-linked peptide detection, there is no ground truth for false positive cross-linked peptides. 
Therefore, researchers developed methods to determine the FDR of detected cross-linked peptides in a data-driven way, even when the ground truth is not known.
Researchers relied on a target-decoy strategy, where verifiably incorrect decoy sequences are appended to the target-sequence database used by the search engine. 
The rate of false positive hits mapping to the target database is then estimated based on the number of hits mapping to the decoy database.
Here, the decoys are conceptually analogous to the knockoffs.
By examining the cross-link detections, four different, equally likely cases can be observed: T-T, T-D, D-T and D-D, where T denotes a target hit and D denotes a decoy hit.
The key idea proposed by \cite{walzthoeni:false} is to use decoy cross-links (\ie T-D, D-T and D-D) to estimate the FDR.

Now we adopt the strategy proposed by \cite{walzthoeni:false} to our setting.
They calculated the expected number of false positive interactions by separately estimating the counts of different types of false interactions.
Specifically, three (not necessarily mutually exclusive) types of false interactions are distinguished below.
Type \textbf{TC--!TC} interactions involve one correct target feature and one feature that is not a correct target feature.
Type \textbf{!TC--!TC} interactions consist of two features, neither of which are correct target features.
Type \textbf{K--K} interactions consist of two features that are both knockoff features (this type can be directly counted).
Note that type \textbf{K--K} interactions are a subset of type \textbf{!TC--!TC} interactions, as knockoff features are not correct target features.

Given a set of detected interactions exceeding a certain score threshold, the total number of false positives $\#(\text{FP})$ in this set can be decomposed as follows:
\begin{equation}
\#(\text{FP})=\#(\text{FP}_\text{TC--!TC}) + \#(\text{FP}_\text{!TC--!TC})
\label{eq:supp:fdr01}
\end{equation}
The expected count of each false positive type given the knockoff counts can be estimated as follows:

\begin{equation}
\begin{aligned}
\widehat{E}\left [ \#(\text{FP}_\text{TC--!TC}) \right ] &= \#(\text{Knockoff}_\text{TC--!TC})\times r_\text{TC--!TC} \\
\widehat{E}\left [ \#(\text{FP}_\text{!TC--!TC}) \right ] &= \#(\text{Knockoff}_\text{!TC--!TC})\times r_\text{!TC--!TC} 
\end{aligned}
\label{eq:supp:fdr02}
\end{equation}
where $\#(\text{Knockoff}_y)$ denotes the number of knockoff-involving interactions of type $y$, and $r_y$ represents the respective original-knockoff frequencies.
For \textbf{TC--!TC} and \textbf{!TC--!TC} interactions we have original-knockoff frequencies $r_\text{TC--!TC}$ of $1:1$ and $r_\text{!TC--!TC}$ of $1:3$.
The original-knockoff frequencies result from the combinatorial composition of the individual types.
For example, the type \textbf{!TC--!TC} comprises four distinct, equally probable cases: O--O, O--K, K--O, and K--K, where O and K denote original and knockoff features, respectively.
The ratio between original-original interactions and knockoff-involving interactions is therefore $1:3$.
In comparison, the type \textbf{TC--!TC} only has two equally probable cases: TC-original and TC-knockoff, which reflects a $1:1$ ratio for this type.

\begin{sloppypar}
However, the number of false positives cannot be determined directly because $\#(\text{Knockoff}_\text{TC--!TC})$ and $\#(\text{Knockoff}_\text{!TC--!TC})$ are not explicitly measurable.
Although knockoff-involving interactions can be identified by the presence of at least one knockoff feature, it is generally not possible to categorize them as either type \textbf{TC--!TC} or \textbf{!TC--!TC}, because original features cannot be definitively classified as correct or incorrect.
Therefore, $\#(\text{Knockoff}_\text{TC--!TC})$ and $\#(\text{Knockoff}_\text{!TC--!TC})$ have to be estimated from the subset of knockoff-involving interactions constituted by two knockoff features (\ie \textbf{K--K}).
Specifically, the expected count of \textbf{!TC--!TC} can be estimated from \textbf{K--K} interactions as follows because the random interactions are distributed equally among the knockoff-involving types O--K, K--O, and K--K:
\begin{equation}
\widehat{E}\left [ \#(\text{Knockoff}_\text{!TC--!TC}) \right ] = 3 \times \#(\text{Knockoff}_\text{K--K})
\label{eq:supp:fdr03}
\end{equation}
Considering that the total number of knockoff-involving interactions \#(\text{Knockoff}) is the sum of $\#(\text{Knockoff}_\text{TC--!TC})$ and $\#(\text{Knockoff}_\text{!TC--!TC})$, it further follows:
\begin{equation}
\widehat{E}\left [ \#(\text{Knockoff}_\text{TC--!TC}) \right ] = \#(\text{Knockoff}) - 3 \times \#(\text{Knockoff}_\text{K--K})
\label{eq:supp:fdr04}
\end{equation}
\end{sloppypar}

The expected number of false positives can now be estimated using the knockoff-involving identification calculations from Eq.\ \ref{eq:supp:fdr03} and Eq.\ \ref{eq:supp:fdr04}:
By plugging these estimates into Eq.\ \ref{eq:supp:fdr02} and using the results to compute the total number of false positives in Eq.\ \ref{eq:supp:fdr01}, we have
\begin{equation}
\widehat{E}\left [ \#(\text{FP}) \right ] = \#(\text{Knockoff}) - 2 \times \#(\text{Knockoff}_\text{K--K})
\label{eq:supp:fdr05}
\end{equation}

\section{Simulated dataset}
\label{sec:supp:simulation}
Following \cite{tsang:detecting}, we use simulation datasets to evaluate \methodname. 
10 simulation functions (described in Tab. \ref{tab:simulation}) are used to generate the simulation datasets.
\begin{table}[h]
  \caption{A test suite of data-generating simulation functions proposed by \cite{tsang:detecting}.}
  \label{tab:simulation}
  \centering
  \begin{tabular}{|c|c|}
  \hline
  \textbf{ID} & \textbf{Simulation Function} \\ \hline
  $F_1$   & $\pi^{x_1 x_2}\sqrt{2x_3}-\sin^{-1}(x_4)+\log(x_3+x_5)-\frac{x_9}{x_{10}}\sqrt{\frac{x_7}{x_8}}-x_2 x_7$         \\ \hline
  $F_2$   & $\pi^{x_1 x_2}\sqrt{2|x_3|}-\sin^{-1}(0.5x_4)+\log(|x_3+x_5|+1)-\frac{x_9}{1+|x_{10}|}\sqrt{\frac{x_7}{1+|x_8|}}-x_2 x_7$         \\ \hline
  $F_3$   & $\exp|x_1-x_2|+|x_2 x_3|- x_3^{2|x_4|}+\log (x_4^2+x_5^2+x_7^2+x_8^2)+x_9+\frac{1}{1+x_{10}^2}$         \\ \hline
  $F_4$   & $\exp|x_1-x_2|+|x_2 x_3|- x_3^{2|x_4|}+(x_1 x_4)^2+\log (x_4^2+x_5^2+x_7^2+x_8^2)+x_9+\frac{1}{1+x_{10}^2}$         \\ \hline
  $F_5$   & $\frac{1}{1+x_1^2+x_2^2+x_3^2}+\sqrt{\exp(x_4+x_5)}+|x_6+x_7|+x_8 x_9 x_{10}$         \\ \hline
  $F_6$   & $\exp(|x_1 x_2|+1)-\exp(|x_3+x_4|+1)+\cos(x_5+x_6-x_8)+\sqrt{x_8^2+x_9^2+x_{10}^2}$         \\ \hline
  $F_7$   & $(\arctan(x_1)+\arctan(x_2))^2+\max(x_3 x_4+x_6,0)-\frac{1}{1+(x_4 x_5 x_6 x_7 x_8)^2}+(\frac{|x_7|}{1+|x_9|})^5 + \sum_{i=1}^{10} x_i$         \\ \hline
  $F_8$   & $x_1 x_2 + 2^{x_3+x_5+x_6}+2^{x_3+x_4+x_5+x_7}+\sin(x_7 \sin(x_8+x_9))+\arccos(0.9x_{10})$         \\ \hline
  $F_9$   & $\tanh (x_1 x_2 + x_3 x_4)\sqrt{|x_5|}+\exp(x_5+x_6)+\log((x_6 x_7 x_8)^2+1)+ x_9 x_{10}+\frac{1}{1+|x_{10}|}$         \\ \hline
  $F_{10}$& $\sinh(x_1+x_2)+\arccos(\tanh(x_3+x_5+x_7))+\cos(x_4+x_5)+\sec(x_7 x_9)$         \\ \hline
  \end{tabular}
\end{table}

Following the settings used in \cite{tsang:detecting}, we employed a sample size of $n=20,000$, equally divided into training and test sets. 
In addition, the number of features is set at $p=30$, and all features are sampled randomly from a continuous uniform distribution, $U(0,1)$.
Only the first $10$ out of $30$ features contribute to the corresponding response, while the remaining features serve as noise to increase the task's complexity.
For robustness, we repeated the experiment 20 times for each simulated dataset using different random seeds. 
Each repetition involved data generation, knockoff generation using KnockoffsDiagnostics \citep{blain:when}, ML model training, and interaction-wise FDR estimation.
For all simulation settings, we reported the mean performance with 95\% confidence intervals, fixing the target FDR level at $q=0.2$.

\begin{figure}[H]
\centering
\includegraphics[width=1.0\textwidth]{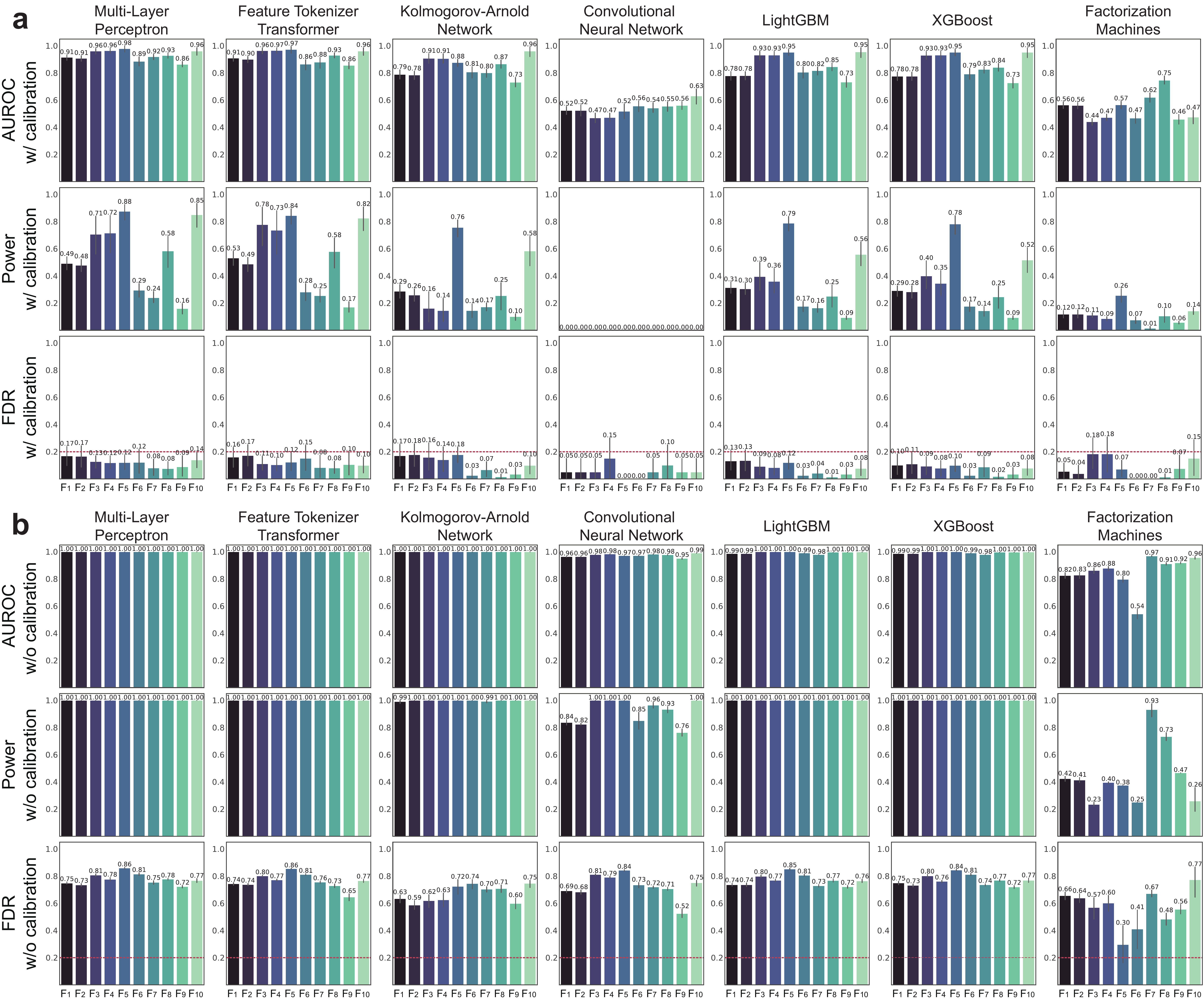}
\caption{
{\bf Evaluation of \methodname\ without non-additivity distillation on simulated datasets.}
(\textbf{a}) \methodname\ with the proposed non-additivity distillation procedure effectively controls the FDR across all ML models.
(\textbf{b}) \methodname\ without the proposed non-additivity distillation procedure fails to control the FDR.
\label{fig:supp:simulation_calib}}
\end{figure}

\begin{figure}[H]
\centering
\includegraphics[width=1.0\textwidth]{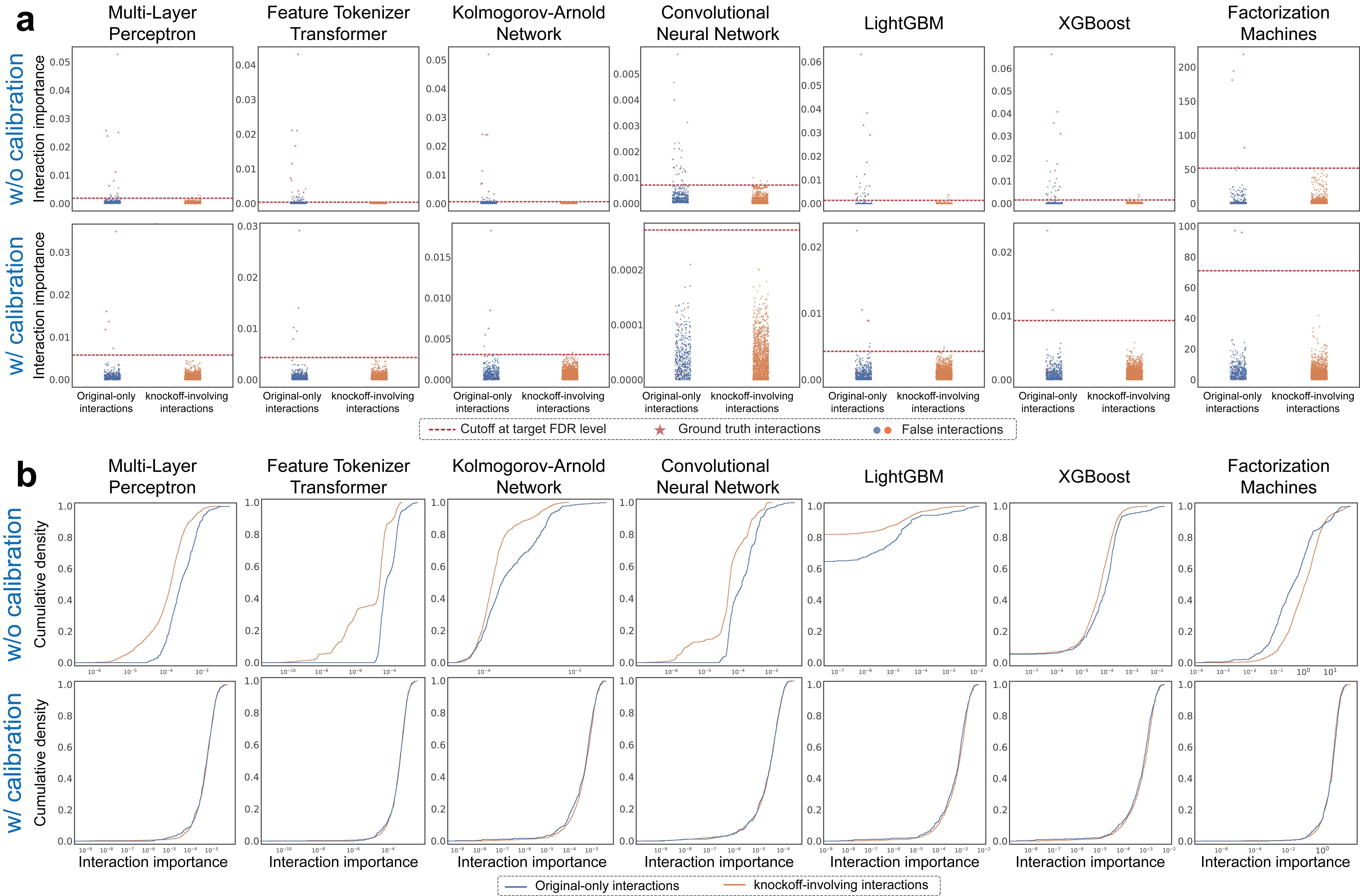}
\caption{
{\bf The non-additivity distillation procedure mitigates the disparity between original-only interactions and those involving knockoffs.}
(\textbf{a}) The reported interaction importance from existing methods reveals a clear disparity in the distribution of interaction importance between original-only interactions and those involving knockoffs, on simulation function $F_1$.
The distilled non-additive interactions help mitigate distributional disparities.
(\textbf{b}) The reported interaction importance from existing methods shows a clear disparity in the cumulative density function between original-only interactions and those involving knockoffs.
\label{fig:supp:simulation_distillation}}
\end{figure}

We discovered that the proposed non-additivity distillation procedure is essential; without it, the FDR cannot be controlled by naively using reported interaction importance from existing methods (Fig. \ref{fig:supp:simulation_calib}).
To gain insight into the FDR control failure, we conducted a qualitative comparison assessing interaction importance before and after non-additivity distillation using the simulation function $F_1$ across different ML models (Fig. \ref{fig:supp:simulation_distillation}). 
The primary cause of the FDR control failure lies in the distribution disparity between original-only interactions and those involving knockoffs.
This suggests a violation of the knockoff filter's assumption in controlling FDR.
The proposed non-additivity distillation procedure mitigates the disparity by extracting non-additive interaction effects from the reported interaction importance measures, thereby enhancing the utility of knockoff-involving interactions as a negative control for FDR estimation.

\section{Real experiments}
\label{sec:supp:real}


\begin{figure}[H]
\centering
\includegraphics[width=1.0\textwidth]{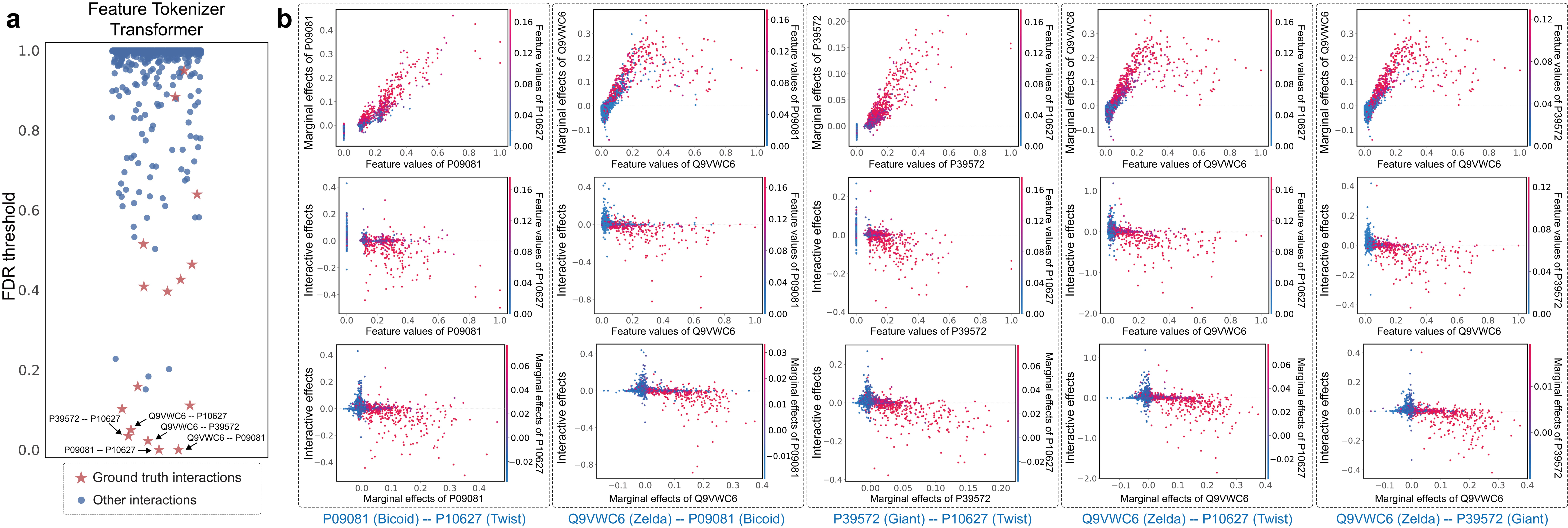}
\caption{
{\bf Evaluating \methodname\ on a real \textit{Drosophila} enhancer dataset with FT-Transformer models.}
(\textbf{a}) \methodname\ is applied with FT-Transformer model to identify important non-additive interactions.
Each possible interaction is measured by the minimum FDR threshold cutoff at which it is selected, with the top five interaction annotated.
The annotated transcription factors are labeled by their UniProt identifiers.
The red stars indicate well-characterized interactions in early \textit{Drosophila} embryos as ground truth. 
(\textbf{b}) The top interactions reported by the MLP model are qualitatively evaluated from three aspects: the contribution of feature values to the marginal and interaction importance measure, and the contribution of the marginal importance measures to the interaction importance measure.
\label{fig:supp:enhancer_transformer}}
\end{figure}

\begin{figure}[H]
\centering
\includegraphics[width=1.0\textwidth]{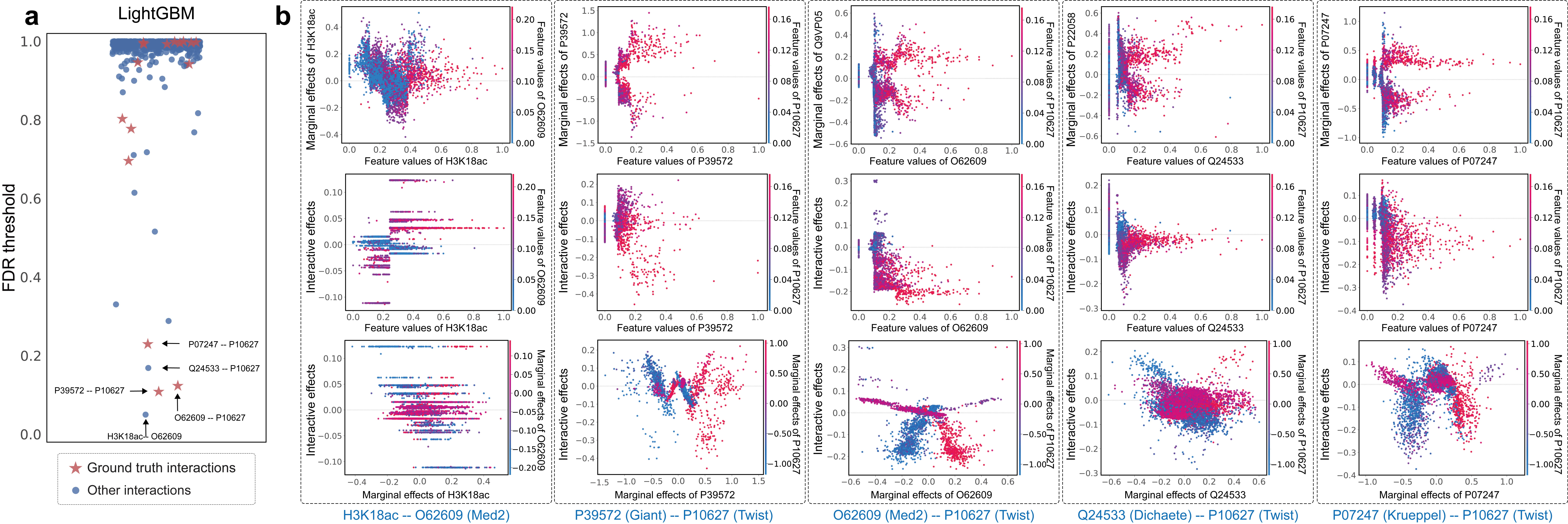}
\caption{
{\bf Evaluating \methodname\ on a real \textit{Drosophila} enhancer dataset with LightGBM models.}
(\textbf{a}) \methodname\ is applied with LightGBM model to identify important non-additive interactions.
Each possible interaction is measured by the minimum FDR threshold cutoff at which it is selected, with the top five interaction annotated.
The annotated transcription factors are labeled by their UniProt identifiers.
The red stars indicate well-characterized interactions in early \textit{Drosophila} embryos as ground truth. 
(\textbf{b}) The top interactions reported by the MLP model are qualitatively evaluated from three aspects: the contribution of feature values to the marginal and interaction importance measure, and the contribution of the marginal importance measures to the interaction importance measure.
\label{fig:supp:enhancer_lightgbm}}
\end{figure}

\begin{figure}[H]
\centering
\includegraphics[width=1.0\textwidth]{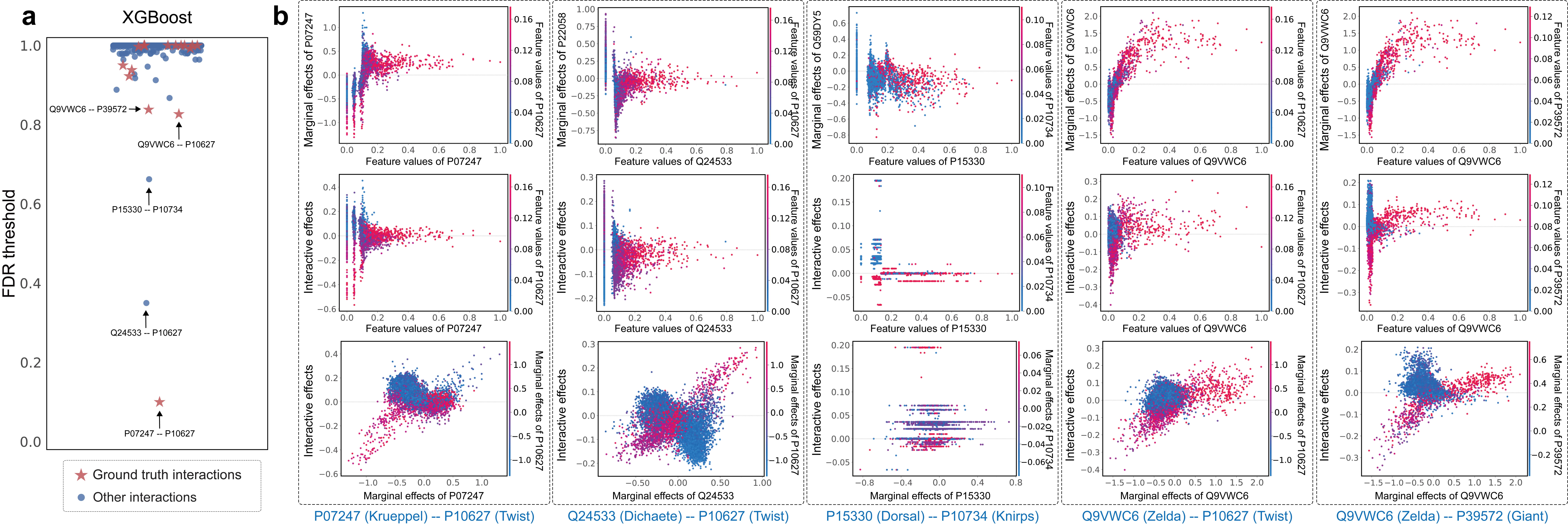}
\caption{
{\bf Evaluating \methodname\ on a real \textit{Drosophila} enhancer dataset with XGBoost models.}
(\textbf{a}) \methodname\ is applied with XGBoost model to identify important non-additive interactions.
Each possible interaction is measured by the minimum FDR threshold cutoff at which it is selected, with the top five interaction annotated.
The annotated transcription factors are labeled by their UniProt identifiers.
The red stars indicate well-characterized interactions in early \textit{Drosophila} embryos as ground truth. 
(\textbf{b}) The top interactions reported by the MLP model are qualitatively evaluated from three aspects: the contribution of feature values to the marginal and interaction importance measure, and the contribution of the marginal importance measures to the interaction importance measure.
\label{fig:supp:enhancer_xgboost}}
\end{figure}

\end{document}